\documentclass[10pt,twocolumn,letterpaper]{article}

\usepackage[pagenumbers]{cvpr}

\usepackage{graphicx}
\usepackage{xspace}
\usepackage{multirow}
\usepackage{booktabs}
\usepackage[table,xcdraw]{xcolor}
\usepackage{soul}
\usepackage{lipsum}
\usepackage{fancyhdr}
\usepackage{pifont}
\usepackage{amsmath}
\usepackage{amssymb}
\usepackage{mathtools}
\usepackage{amsthm}
\usepackage{colortbl}
\usepackage{adjustbox}
\usepackage[ruled]{algorithm2e}
\usepackage{upgreek}
\usepackage{wrapfig}
\usepackage{tcolorbox}
\usepackage{textgreek}
\usepackage{bbm}
\usepackage{cuted}
\usepackage{capt-of}

\definecolor{cvprblue}{rgb}{0.21,0.49,0.74}
\definecolor{greenx}{RGB}{0,128,128}
\definecolor{maroonx}{RGB}{195,18,48}
\definecolor{darkred}{RGB}{192,0,0}
\definecolor{darkgreen}{RGB}{103,174,64}

\usepackage[
    colorlinks=true,
    linkcolor=maroonx,
    anchorcolor=blue,
    pagebackref,
    citecolor=cvprblue
]{hyperref}

\newcommand{\cmark}{\ding{51}}
\newcommand{\xmark}{\ding{55}}

\newcommand{\cc}{\cellcolor{gray!20}}
\newcommand{\name}{\texttt{Kestrel}\xspace}
\newcommand{\RebuttalRevision}[1]{\textcolor{black}{#1}}

\theoremstyle{definition}

\def\paperID{8128}
\def\confName{CVPR}
\def\confYear{2026}

\newcommand{\kestrelicon}{%
  \makebox[0pt][r]{%
    \raisebox{-0.62em}{\includegraphics[height=1.9em]{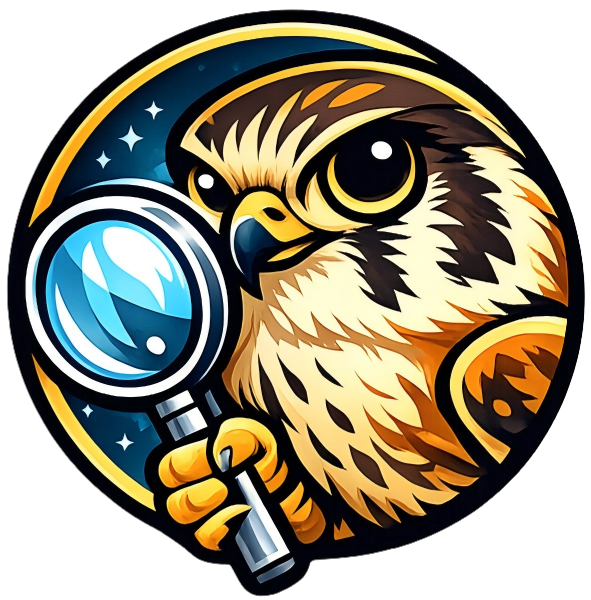}}%
    \kern0.28em
  }%
}

\title{\texorpdfstring{%
  \makebox[\textwidth][c]{%
    \kestrelicon
    \begin{tabular}[t]{@{}c@{}}
      Kestrel: Grounding Self-Refinement for 
      LVLM Hallucination Mitigation
    \end{tabular}%
  }%
}{%
  Kestrel: Grounding Self-Refinement for LVLM Hallucination Mitigation%
}}

\author{%
Jiawei Mao$^{1}$
\quad
Hardy Chen$^{1}$ \quad
Haoqin Tu$^{1}$ \quad
Yuhan Wang$^{1}$ \quad
Letian Zhang$^{1}$ \quad 
Zeyu Zheng$^2$ \\
Huaxiu Yao$^3$ \quad
Zirui Wang$^4$ \quad
Cihang Xie$^1$ \quad
Yuyin Zhou$^1$ \vspace{1.5em} \\
$^1$UC Santa Cruz
\quad $^2$UC Berkeley \quad $^3$UNC-Chapel Hill \quad $^4$Apple
}

\begin{document}
\maketitle

\vspace{-2.8em}
\begin{strip}
    \centering
    \vspace{-3.5em}
    \includegraphics[width=\textwidth]{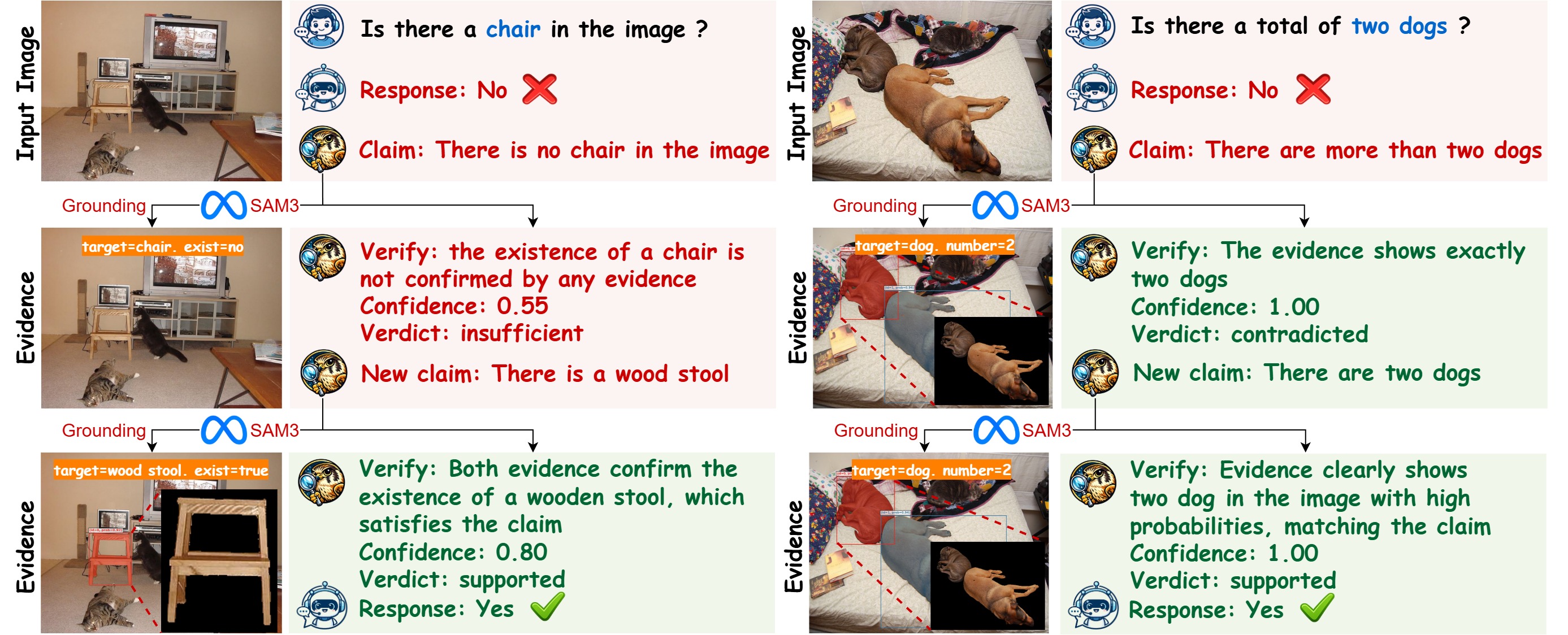}
    \captionof{figure}{
    \textbf{\name} progressively corrects hallucinated LVLM answers by integrating an external grounding agent with iterative self-improvement. At each round, the model grounds the current claim with explicit visual and textual evidence, conducts claim-level verification, and conservatively refines the response, yielding a final answer that is both more reliable and more interpretable.
    }
    \label{fig_teaser}
    \vspace{-0.8em}
\end{strip}

\begin{abstract}
Large vision-language models (LVLMs) have become increasingly strong but remain prone to hallucinations in multimodal tasks, which significantly narrows their deployment. As training these LVLMs to avoid hallucinations becomes prohibitively expensive for larger models, training-free methods offer a cheap and flexible solution to this problem, yet existing approaches based on decoding or tool use often bring limited gains and/or weak interpretability.
We propose \textbf{\name}, a training-free framework for LVLM hallucination mitigation that combines an explicit visual-grounding agent with evidence-verified self-refinement mechanism. 
In detail, \name\ first collects explicit visual evidence and converts tool outputs into reusable and structured textual evidence. Second, to take full advantage of these evidence, \name\ verifies them via an LVLM judge for evidence checking, then iteratively self-refine answers based on verified evidence to reduce the risk of over-correction.
Extensive experiments show that \name\ improves performance over strong baselines across hallucination benchmarks (\eg, average +3.31$\%$ on POPE and +28.34 on MME-Hallucination with Qwen3-VL), while providing transparent verification traces for hallucination diagnosis and analysis --- \eg, both the integrated self-refinement module and grounding agent contributing an average +2.0$\%$ gain on POPE. Project website: \url{https://jwmao1.github.io/Kestrel_project/}
\end{abstract}
\section{Introduction}
\label{sec:intro}

Recent advances in large-scale pretraining~\cite{radford2021learning,alayrac2022flamingo,zhai2023sigmoid} and multimodal instruction tuning~\cite{liu2023visual,dai2023instructblip} have substantially improved the capabilities of large vision-language models (LVLMs)~\cite{grattafiori2024llama,bai2025qwen3,wang2025internvl3} on multimodal understanding and reasoning tasks such as visual question answering (VQA). 
However, LVLMs still exhibit hallucination, producing responses that are inconsistent with or weakly supported by the input image. 
For example, empirical studies~\cite{li2023evaluating,wang2023amber,sun2024aligning} show that this issue remains prevalent, making hallucination a central challenge for improving the reliability of LVLMs.

To mitigate hallucination, two broad classes of methods have been proposed, training-based and training-free. For the training-based line of work, continual training with hallucination annotations or alignment with external feedback has been shown effective~\cite{zhang2024reflective,chen2023mitigating,jiang2024hallucination,sun2024aligning,lyu2024alleviating}. 
However, these training-based solutions incur significant data and compute overhead, posing hurdles in real-world deployment.
Existing training-free methods for hallucination mitigation improve test-time correction without additional training, but leave key gaps to be filled: (i) limited gains and robustness when operating purely on internal decoding dynamics without external grounding evidence, and (ii) limited reliability when correction is performed in a single pass. 
Distribution-contrast methods~\cite{damonlpsg2023vcd,woo2024ritual} can reduce object hallucinations but remain sensitive to perturbations and often favor common-object representations. Many approaches rely on internal logit dynamics~\cite{huang2025shield} or language-level decoding control~\cite{huang2024opera}, which can yield brittle corrections that are difficult to validate against concrete visual evidence. 
On the other hand, methods that introduce external verification may produce non-deterministic evidence due to randomness of tools~\cite{zhang2025self}. 
While other methods~\cite{yin2024woodpecker} could collect reliable evidence, their one-time verification-and-update can be insufficient to prevent over-correction under challenging cases.


\begin{figure*}[h!]
    \centering
    \includegraphics[width=1\linewidth]{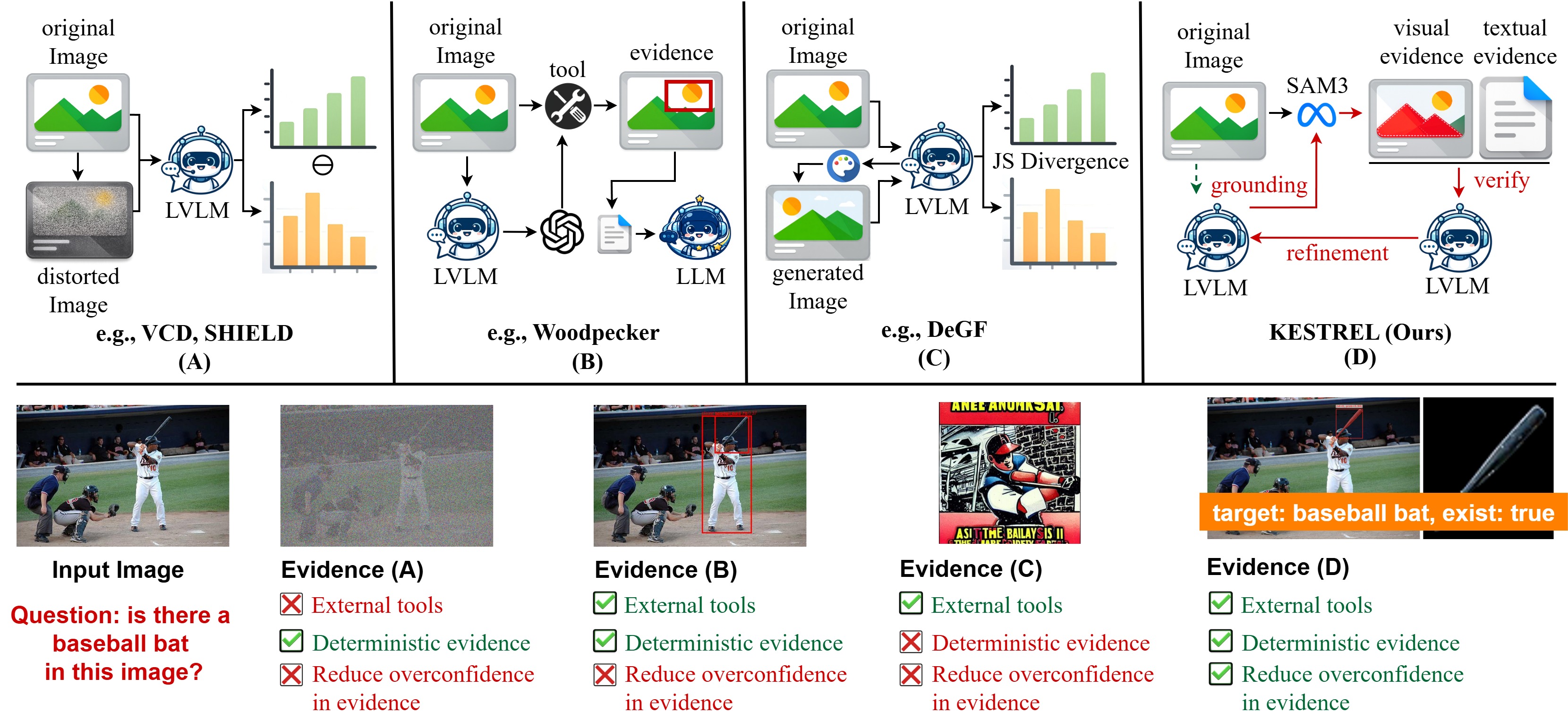}
     \caption{
     \textbf{\name vs. prior training-free hallucination mitigation methods.} By combining an external grounding agent with iterative self-improvement, \name collects explicit visual evidence and further converts tool outputs into structured textual evidence for verification. This design yields more interpretable and stable evidence, reduces overconfident corrections, and avoids biased interpretation that may arise when LVLMs rely only on raw visual evidence compared with prior approaches.
     }
     \label{fig_paradigm}
   \end{figure*}

Building on these limitations, we propose \name (see Fig.~\ref{fig_teaser}), a training-free framework for LVLM hallucination mitigation that unifies an explicit visual grounding agent with evidence-driven iterative self-refinement. 
Specifically, \name first decomposes the question into verifiable claim-level targets (e.g., existence, color, count, and position), and then invokes SAM3~\cite{carion2025sam3segmentconcepts} around each target to collect segmentation overlays, bounding boxes, target crop-and-zoom views, and text evidence derived from these collected visual evidence, all of which are collated as structured evidence items with citation identifiers. 
The framework then performs claim-level verification with verdicts and outputs confidence-aware verification results, forming an auditable evidence chain. 
To regulate the potential over-correction, we further introduce an evidence-gated update scheme into the iteration: the framework progressively supplements and strengthens claim-level evidence through multiple rounds of verification and revision, and permits answer flips only when evidence strength, confidence, and evidence coverage jointly satisfy predefined criteria. 
These designs preserve the training-free methods while improving the interpretability, robustness, and decision stability of hallucination mitigation.


Experiments show that \name remains fully training-free, yet consistently reduces hallucinations at test time across multiple benchmarks, with improvements that transfer across different state-of-the-art LVLM backbones. On POPE~\cite{li2023evaluating} (MS-COCO, A-OKVQA, and GQA), \name improves accuracy by an average of +3.31$\%$ points over Qwen3-VL and +3.03$\%$ over InternVL3.5; it also surpasses prior training-free mitigation baselines by +1.38$\%$ and +1.47$\%$ points on average under the same backbones, respectively. On the more challenging MME-Hallucination~\cite{fu2023mme}, \name boosts Qwen3-VL by +28.34 points and exceeds OPERA~\cite{huang2024opera} by +16.67, delivering consistent gains across diverse hallucination types (existence, count, and position) while maintaining strong overall performance and setting a new state-of-the-art.

Our main contributions are summarized as follows:

$\bullet$ We propose \name, a training-free LVLM hallucination mitigation framework that unifies explicit visual grounding agent with iterative self-refinement at test time. \name decomposes answers into verifiable claim-level targets, grounds them with structured visual and textual evidence, performing conservative multi-round verification and revision to improve interpretability and reduce over-correction.

$\bullet$ \name achieves state-of-the-art performance in hallucination mitigation on POPE and the more fine-grained MME-Hallucination.

$\bullet$ \name generalizes across multiple state-of-the-art LVLM backbones with substantial and consistent gains, showing that the framework is backbone-agnostic and broadly applicable in the training-free setting.
\section{Related Work}
\label{sec:related_work}

\subsection{Large Vision-Language Models}
Large vision-language models (LVLMs) have advanced rapidly through large-scale multimodal pretraining and instruction tuning, achieving strong performance across multimodal understanding and reasoning tasks. Representative paradigms include CLIP-style vision-language pretraining~\cite{radford2021learning}, Flamingo-style few-shot multimodal modeling~\cite{alayrac2022flamingo}, and BLIP-2 style~\cite{li2023blip} modular alignment between frozen vision encoders and LLMs. LVLMs, such as LLaVA~\cite{liu2023visual,liu2024improved}, InstructBLIP~\cite{dai2023instructblip}, OpenFlamingo~\cite{awadalla2023openflamingo}, CogVLM~\cite{wang2024cogvlm}, Kosmos-2~\cite{peng2023kosmos}, and recent models~\cite{bai2025qwen3,wang2025internvl3,grattafiori2024llama,chen2024allavaharnessinggpt4vsynthesizeddata,chen2023sharegpt4vimprovinglargemultimodal}, further demonstrate the effectiveness of scalable multimodal alignment and visual instruction tuning. Meanwhile, grounded multimodal modeling has become increasingly important, as exemplified by Kosmos-2~\cite{peng2023kosmos}, which explicitly supports phrase grounding and visual referring. Nevertheless, current LVLMs still struggle to maintain faithful grounding between generated responses and image content, especially in fine-grained reasoning scenarios, making hallucination a persistent challenge for reliable deployment.

\subsection{Hallucination in LVLMs}
Hallucination is a persistent problem in large vision-language models (LVLMs). Early studies~\cite{li2023evaluating} show that LVLMs often generate content inconsistent with the input image, especially by predicting non-existent objects, while POPE~\cite{li2023evaluating} improves the stability of such evaluation. Subsequent work shows that hallucination extends beyond object existence to finer-grained errors in attributes, counts, and relations, as benchmarked by AMBER~\cite{wang2023amber}. More challenging settings, such as visual illusion and ambiguous local evidence, are further explored in HallusionBench~\cite{guan2024hallusionbench}. Broader benchmarks including MME~\cite{fu2023mme}, MMHal-Bench~\cite{sun2024aligning}, and THRONE~\cite{kaul2024throne} further suggest that hallucination is heterogeneous, benchmark-sensitive, and closely tied to failures in visual grounding and multimodal reasoning. These findings motivate mitigation methods that verify model outputs against explicit and fine-grained visual evidence.

\subsection{Training-based Hallucination Mitigation}
Early approaches improve faithfulness by redesigning instruction data or supervision signals so that models better distinguish grounded from ungrounded responses. For example, robust visual instruction tuning~\cite{chen2023mitigating} introduces hallucination-oriented supervision, while HACL~\cite{jiang2024hallucination} uses contrastive learning to separate grounded and hallucinated representations. Reflective instruction tuning~\cite{zhang2024reflective} further improves reliability by adding rationale supervision. Alignment-based methods, such as factually augmented RLHF~\cite{sun2024aligning} and Silkie~\cite{li2023silkie}, incorporate preference or factual signals during post-training, and HIO~\cite{lyu2024alleviating} strengthens token-level contrastive learning around hallucinated content. Overall, training-based methods are effective, but they usually require additional annotations, synthetic data, preference collection, or repeated optimization, leading to higher training cost and deployment complexity.

\begin{figure*}[h!]
    \centering
    \includegraphics[width=1\linewidth]{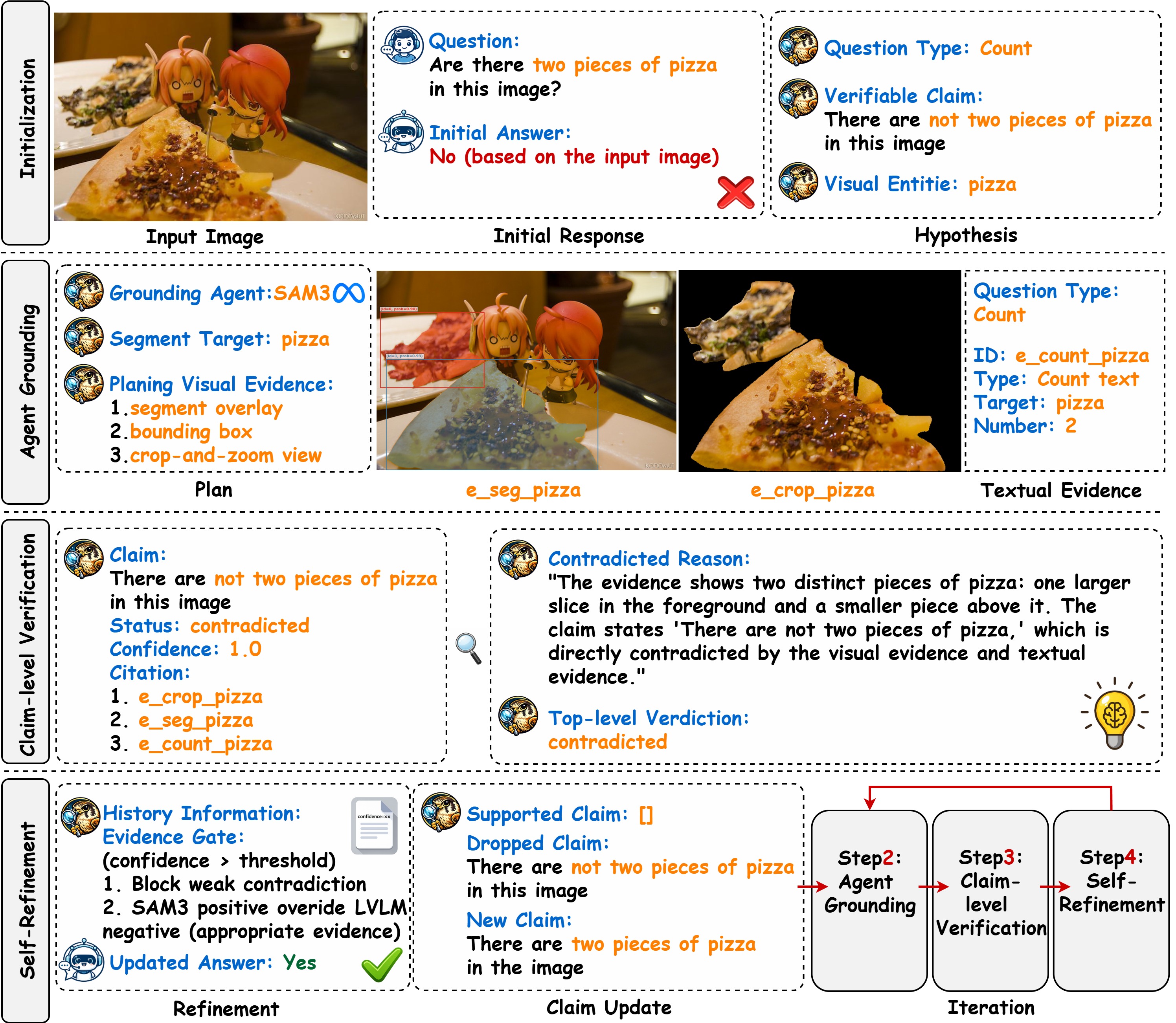}
     \caption{
     \textbf{Overview of \name.} Given an image-question pair, \name follows a training-free four-stage pipeline for LVLM hallucination mitigation: \textbf{(1) Initialization}, which obtains an initial answer and rewrites it into question-aligned verifiable claims with associated visual entities and claim types; \textbf{(2) Agent Grounding}, which invokes an external SAM3-based grounding agent to collect explicit visual evidence (e.g., segmentation overlays, boxes, and crop-and-zoom views) and convert them into structured textual evidence; \textbf{(3) Claim-level Verification}, which verifies each claim against the cited evidence to produce claim-wise verdicts, confidence scores, and a top-level verification decision; and \textbf{(4) Self-Refinement}, which performs evidence-gated answer updating based on the current and previous verification traces. 
     }
     \vspace{-0.2cm}
     \label{fig_method}
   \end{figure*}

\subsection{Training-free Hallucination Mitigation}
Training-free hallucination mitigation aims to reduce hallucination at inference time without updating model parameters. A major line of work focuses on contrastive or decoding-based strategies, such as VCD~\cite{damonlpsg2023vcd}, RITUAL~\cite{woo2024ritual}, OPERA~\cite{huang2024opera}, and SHIELD~\cite{huang2025shield}, which alleviate hallucination by intervening on decoding behavior or visual token representations. Another line introduces explicit verification or post-hoc correction. For example, Woodpecker~\cite{yin2024woodpecker} adopts a multi-stage correction pipeline, while DeGF~\cite{zhang2025self} leverages text-to-image generative feedback for iterative refinement. Meanwhile, recent grounding models such as SAM3~\cite{carion2025sam3segmentconcepts} make it increasingly practical to collect explicit visual evidence at inference time. Compared with prior training-free methods, our work further emphasizes combine explicit grounding evidence and conservative iterative self-refinement to mitigate hallucination.
\section{Method}
\label{sec:method}

We propose \name, a training-free framework for mitigating LVLM hallucination with explicit visual grounding agent and structured evidence-driven self-refinement at test time.
Given an image $\mathcal{I}$ and corresponding question $\mathcal{Q}$, \name iteratively follows a four-step pipeline:
\textbf{(i)} initialization,
\textbf{(ii)} agent grounding,
\textbf{(iii)} claim-level verification, and
\textbf{(iv)} self-refinement (see Fig.~\ref{fig_method}).

\subsection{Initialization}
\label{sec:init}
\name first queries the LVLM to obtain an initial answer $\hat{\mathcal{A}}^{(0)}$.
To support claim-level verification, \name converts $\mathcal{Q}$ into a small set of \emph{verifiable claims} that directly correspond to the question. 
Concretely, \name rewrites the question-answer decision into visually checkable claims, each anchored to one or two concrete visual entities. 
These entities serve as the detection targets for the grounding agent. 
Meanwhile, based on the verifiable attributes required by the question, we categorize the extracted claims (e.g., \texttt{existence}, \texttt{color}, \texttt{count}, \texttt{position}) to route subsequent agent grounding. 

\subsection{Agent Grounding}
\label{sec:tools}
To obtain explicit, inspectable grounding evidence, \name invokes an external visual grounding agent built on \textbf{SAM3} promptable concept segmentation~\cite{carion2025sam3segmentconcepts}.  

\paragraph{Visual evidence.}
SAM3 takes the visual entities in the claims as concept prompts and returns the matched instances, from which \name collects explicit visual evidence, including:
\textbf{(i)} segmentation overlays for transparent localization,
\textbf{(ii)} instance bounding boxes (derived from SAM3 masks) to support geometry-based reasoning,
and \textbf{(iii)} crop-and-zoom views around predicted instances to reduce ambiguity for attribute inspection (e.g., color) and local details.

\paragraph{Structured textual evidence.}
To make agent outputs directly usable for claim verification and auditable diagnosis, \name derives textual evidence from visual evidences via LVLM for each claim type:  
\textbf{(i)} for \texttt{existence}, we convert the predicted instances into an existence statement by checking whether the number of matched instances is greater than zero; 
\textbf{(ii)} for \texttt{count}, we report the instance count computed from the number of predicted masks; 
\textbf{(iii)} for \texttt{color}, we generate a concise color observation conditioned on the masked crop-and-zoom and the full image; 
\textbf{(iv)} for \texttt{position}, we convert SAM3 geometry into text by deriving coarse spatial cues from the union bounding box and, when two entities are involved, computing their relative relation from the corresponding bounding-box centers. 
Each textual evidence item is paired with a citation identifier and can be referenced during verification and answer revision.

\subsection{Claim-level Verification}
\label{sec:verify}
Given the claims and corresponding structured evidence items, \name performs \textbf{claim-level verification} using LVLM-as-a-judge. 
The judge is instructed to base its decision only on the provided evidence and to cite the corresponding evidence. 
For each claim, the verifier outputs:
\textbf{(i)} a verdict (\emph{supported} / \emph{contradicted} / \emph{insufficient}),
\textbf{(ii)} a confidence score,
and \textbf{(iii)} a short reasoning that must cite the relevant evidence. 
We then consolidate claim-wise judgments into a top-level verification verdict for the current answer: it is labeled as \emph{contradicted} if any claim is confidently refuted with cited evidence, as \emph{supported} only when all claims are confidently supported, and as \emph{insufficient} otherwise. The resulting verification trace constitutes an explicit, evidence-grounded audit trail, enabling interpretable analysis of when hallucinations arise and how corrections are triggered.

\subsection{Self-Refinement}
\label{sec:update}
Since external agent and the LVLM may be unrobust, directly revising the answer based on the instance-level verdict can introduce \emph{over-correction}. Therefore, \name adopts a evidence-gated self-refinement strategy: it permits correction {only when} the verification provides sufficiently reliable signals---i.e., high-confidence claim-level judgments together with cited evidence for the corresponding claim. Otherwise, \name preserves the current answer $\hat{\mathcal{A}}^{(i)}$ (where $i$ denotes $i$ th iteration) and proceeds to collect stronger evidence in subsequent rounds. 

Importantly, the self-refinement is \textbf{stateful}: the revision step conditions not only on the current verification results but also on prior rounds' claims, evidence, and decisions. Based on the verification trace, \name updates the answer to obtain $\hat{\mathcal{A}}^{(i+1)}$ and proposes a new set of claims for the next iteration, prioritizing claims that remain uncertain or are implicated by contradictions. This iterative process progressively strengthens evidence and stabilizes decision-making, while remaining training-free.

\name repeats the cycle for a small number of iterations, and stops early when the answer stabilizes under consistently supportive verification, or when additional iterations no longer yield stronger evidence.
The final output is the answer together with its claim-level verification traces.

\section{Experiments}
\label{sec:experiments}

\subsection{Experimental Setup}

\begin{table*}[t]
    \renewcommand{\arraystretch}{1.05}
    \centering
    \small
    \caption{
        \RebuttalRevision{\textbf{Results on POPE~\citep{li2023evaluating} benchmark}. Higher ($\uparrow$) accuracy indicates better performance. The best results are \textbf{bolded}, and the second-best are \underline{underlined}.}
    }
    \label{tab:POPE}
    \setlength{\tabcolsep}{7pt} 
    \resizebox{\textwidth}{!}{
    \begin{tabular}{clccccccccc}
    \toprule
     \multirow{2}{*}[-2pt]{\textbf{Backbone}} & \multirow{2}{*}[-2pt]{\textbf{Method}} & \multicolumn{3}{c}{\textbf{MS-COCO}~\cite{lin2014microsoft}} & \multicolumn{3}{c}{\textbf{A-OKVQA}~\cite{schwenk2022okvqa}} & \multicolumn{3}{c}{\RebuttalRevision{\textbf{GQA}}~\cite{hudson2019gqa}} \\
    \arrayrulecolor{gray} \cmidrule(lr){3-5} \cmidrule(lr){6-8} \cmidrule(lr){9-11}
     &  & {Rand.} $\uparrow$ & {Pop.}  $\uparrow$ & {Adv.} $\uparrow$ & {Rand.} $\uparrow$ & {Pop.}  $\uparrow$ & {Adv.} $\uparrow$ & \RebuttalRevision{{Rand.} $\uparrow$} & \RebuttalRevision{{Pop.} $\uparrow$} & \RebuttalRevision{{Adv.} $\uparrow$} \\
    \midrule
    \multirow{8}{*}[-5pt]{\rotatebox{90}{\textbf{Qwen3-VL}~\cite{bai2025qwen3}}}  
    & Baseline~\cite{bai2025qwen3} & 89.00 & 86.92  & 86.20 & 92.36 & 86.67  & 81.87 & 90.70 & 83.63 & 81.50 \\
    & Qwen3-VL agent~\cite{bai2025qwen3} & \underline{91.03} & 88.06  & 86.13 & 92.87 & 85.20  & 78.03 & 91.41 & 81.81 & 78.30 \\ \cmidrule(lr){2-11}
     & VCD~\cite{damonlpsg2023vcd}  & 90.40 & 88.80  & 87.41 & \underline{93.53} & 87.86  & 82.00 & 91.56 & 85.76 & 81.93 \\
     & Woodpecker~\cite{yin2024woodpecker}  &  89.97 &  88.03  &  87.10 &  93.23 &  88.90  &  83.33 & 91.27 & 86.27 & 82.77  \\
     & RITUAL~\cite{woo2024ritual}  &  86.20	&  83.67 &   82.27 &  87.67 &  83.50  &  77.76 & 86.86 & 82.30 & 78.23 \\ 
     & OPERA~\cite{huang2024opera}  &  90.50	&  \underline{88.83} &   \underline{87.50} &  \textbf{93.76} &  \underline{89.50}  &  \underline{83.86} & \textbf{91.80} & \underline{87.11} & \underline{83.30} \\ 
     & DeGF~\cite{zhang2025self}  &  90.33	&  88.16 &   86.90 &  92.96 &  87.70  &  82.61 & 91.13 & 83.79 & 82.00 \\
     & {\textbf{\name}}$_{\mathrm{(ours)}}$ &\cc \textbf{91.53} &\cc \textbf{89.30}  &\cc \textbf{87.53}  &\cc 93.46  &\cc \textbf{91.73}  &\cc \textbf{86.76}  & \cc \underline{91.67} & \cc \textbf{90.33} & \cc \textbf{86.27} \\
     \arrayrulecolor{gray}\cmidrule(lr){1-11}
     \multirow{7}{*}[-5pt]{\rotatebox{90}{\textbf{ InternVL3.5}~\cite{wang2025internvl3}}} & Baseline~\cite{wang2025internvl3} & 90.77 & 88.10  & 85.73 & 92.67 & 87.83  & 81.53 & 89.77 & 84.10 & 81.31 \\ \cmidrule(lr){2-11}
     & VCD~\cite{damonlpsg2023vcd}  & 91.35 & \underline{89.22}  & \underline{87.60} & 92.87 &  89.73  & 83.73 & \textbf{91.60} & 85.07 & \underline{83.37} \\
     & Woodpecker~\cite{yin2024woodpecker}  &  91.20 &  89.11  &  87.50 &  \textbf{93.73} & \underline{89.80}  &  84.00 & 91.43 & 85.16 & 83.26 \\
     & RITUAL~\cite{woo2024ritual}  &  \textbf{91.60}	&  89.03 &   87.48 &  \underline{93.71} &  89.75  &  83.90 & 91.39 & 85.18 & 83.29 \\
     & OPERA~\cite{huang2024opera}  &  \underline{91.53}	&  89.18 &   87.55 &  93.55 &  89.79  &  83.81 & 91.45 & \underline{85.20} & 83.31 \\ 
     & DeGF~\cite{zhang2025self}  &  91.43	&  89.12 &   87.37 &  93.39 &  89.68  &  \underline{84.11} & 91.48 & 85.06 & 83.20 \\
     & {\textbf{\name}}$_{\mathrm{(ours)}}$ &\cc 91.27 &\cc \textbf{89.27}  &\cc \textbf{88.10} &\cc 93.57 &\cc \textbf{91.80}  &\cc \textbf{87.13} & \cc \underline{91.57} & \cc \textbf{89.87} & \cc \textbf{86.53} \\
     \arrayrulecolor{gray}\cmidrule(lr){1-11}
    \bottomrule
    \end{tabular}
    }
\end{table*}

\paragraph{Benchmarks.}
We evaluate \name on \textbf{POPE}~\cite{li2023evaluating} in three source datasets (MS-COCO~\cite{lin2014microsoft}, A-OKVQA~\cite{schwenk2022okvqa}, and GQA~\cite{hudson2019gqa}) under random (Rand.), popular (Pop.), and adversarial (Adv.) sampling, and \textbf{MME-Hallucination}~\cite{fu2023mme} evaluates fine-grained hallucination across existence, count, position, and color. 

\paragraph{LVLM backbones.}
We evaluate \name with SoTA open-weight LVLMs, e.g.,
\textbf{Qwen3-VL 8B}~\cite{bai2025qwen3} and \textbf{InternVL3.5 8B}~\cite{wang2025internvl3}.

\paragraph{Baselines.}
We compare against Qwen3-VL agent (with zoom-in tool)\footnote{\url{https://github.com/QwenLM/Qwen-Agent/blob/main/examples/cookbook_think_with_images.ipynb}} and training-free LVLM hallucination mitigation baselines: \textbf{VCD}~\cite{damonlpsg2023vcd}, \textbf{OPERA}~\cite{huang2024opera}, \textbf{RITUAL}~\cite{woo2024ritual},  \textbf{Woodpecker}~\cite{yin2024woodpecker}, and \textbf{DeGF}~\cite{zhang2025self}.
All baselines use the same LVLM backbone for fair comparison (In the Woodpecker pipeline, all components except the visual verification module are replaced with Qwen3-VL.).

\paragraph{Implementation details.}
We set the maximum number of self-refinement iterations to $K{=}3$ and stop early when the answer is stable with two consecutive \textit{supported} verification verdicts.
For the SAM3 grounding agent, we use a confidence threshold of $0.5$ (with a recheck threshold of $0.35$ when needed for existence claim).
In self-refinement, we use confidence thresholds in the range $[0.82,\,0.90]$ for different claim types.
All experiments are conducted on NVIDIA A100 GPUs. Details of the corresponding prompt template are provided in the supplementary material.

\subsection{Results And Discussion}

\paragraph{Results on POPE.} 
Tab.~\ref{tab:POPE} compares \name with baselines under two LVLM backbones (Qwen3-VL and InternVL3.5) across three evaluation sources (MS-COCO, A-OKVQA, and GQA).
Overall, \name exhibits the most consistent gains on the more challenging popular and adversarial splits, while remaining competitive on random sets, indicating robust hallucination mitigation across backbones and data sources.

With \textbf{Qwen3-VL}, \name achieves the best performance on \textbf{MS-COCO} across all splits.
On \textbf{A-OKVQA}, \name reaches \textbf{91.73$\%$} (Pop.) and \textbf{86.76$\%$} (Adv.), surpassing decoding-based baselines such as OPERA (89.50$\%$/83.86$\%$) and VCD (87.86$\%$/82.00$\%$).
On \textbf{GQA}, \name further improves to \textbf{90.33$\%$} (Pop.) and \textbf{86.27$\%$} (Adv.), outperforming OPERA (87.11$\%$/83.30$\%$) and VCD (85.76$\%$/81.93$\%$) by a clear margin.
With \textbf{InternVL3.5}, we observe a similar trend, delivering the largest improvements over the LVLM backbone and the strongest overall performance.

\begin{figure*}[h!]
    \centering
    \includegraphics[width=1\linewidth]{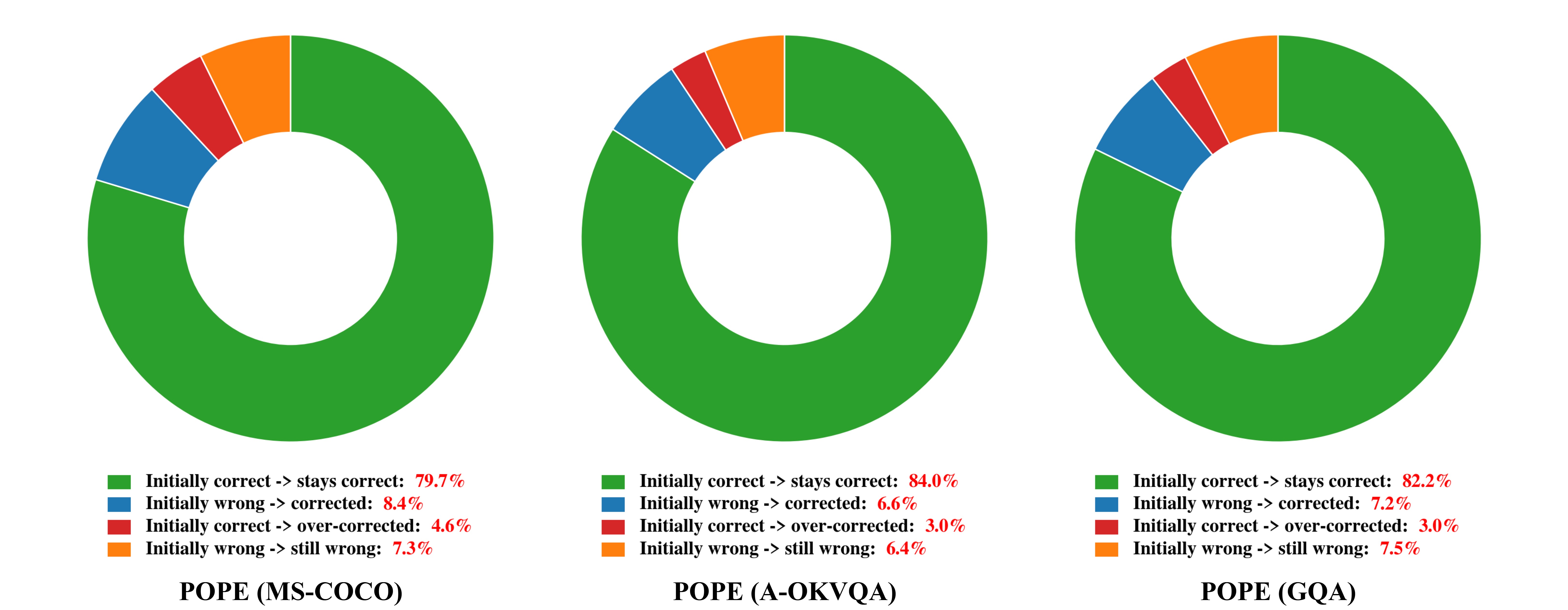}
     \caption{
     \textbf{Prediction transition statistics with Qwen3-VL before and after refinement.} Prediction are categorized into four types: correctly preserved, error corrected, over-corrected, and incorrectly preserved. The results show that the refinement process is conservative, retaining most originally correct predictions while correcting a portion of erroneous ones, with limited over-correction. Zoom in for a better view.
     }
     \label{fig_bar}
   \end{figure*}

Compared to decoding-centric methods that reshape token distributions at inference time (e.g., VCD~\cite{damonlpsg2023vcd} and OPERA~\cite{huang2024opera})
and post-hoc correction pipelines such as Woodpecker~\cite{yin2024woodpecker},
\name externalizes grounding into explicit evidence and performs claim-level verification with conservative, evidence-gated updates.
This design better preserves correctness while reducing risky corrections under stronger prior interference. 
Fig.~\ref{fig_bar} further shows that the gains on POPE mainly come from correcting initially incorrect predictions, while most correct predictions are preserved after refinement. The relatively low rate of over-correction suggests that our conservative gating mechanism effectively suppresses unnecessary revisions, enabling the model to mitigate hallucinations without sacrificing prediction stability.

\begin{figure*}[h!]
    \centering
    \includegraphics[width=1\linewidth]{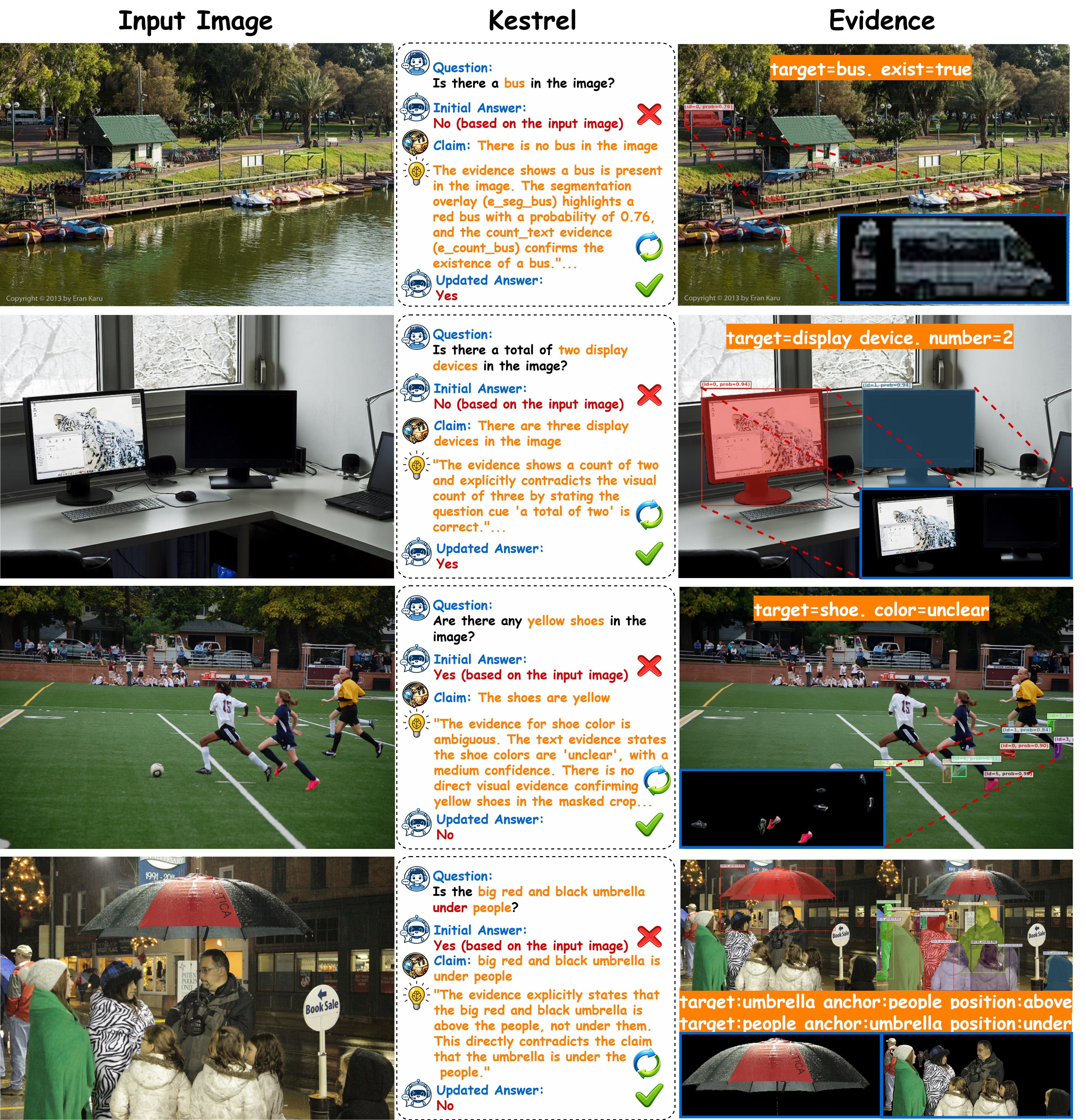}
     \caption{
     \textbf{Qualitative results of \name.} We compare the VQA responses from the regular baseline and our method based on Qwen3-VL. Zoom in for a better view.
     }
     \label{fig_exp}
     \vspace{-0.2cm}
   \end{figure*}

\paragraph{Results on MME-Hallucination.}

\begin{table*}[t!]
        \begin{center}
        \begin{small}
        \setlength{\tabcolsep}{9pt} 
        \caption{\looseness=-1\textbf{Results on MME-Hallucination~\citep{fu2023mme} benchmark.} Higher scores ($\uparrow$) indicate better performance. The best results are \textbf{bolded}, and the second-best are \underline{underlined}. }
        \label{tab:MME}
        \resizebox{\textwidth}{!}{
        \begin{tabular}{lcccccc}
            \toprule
             \multirow{2}{*}[-0.5ex]{\textbf{Backbone}}  & \multirow{2}{*}[-0.5ex]{\textbf{Method}}  &  \multicolumn{2}{c}{\textbf{Object-level}} & \multicolumn{2}{c}{\textbf{Attribute-level}} & \multirow{2}{*}[-0.5ex]{\textbf{MME Score $\uparrow$}} \\
             \cmidrule(lr){3-4}\cmidrule(lr){5-6}
                & & Existence $\uparrow$ & Count $\uparrow$ & Position $\uparrow$ & Color $\uparrow$   & \\
             \midrule
            \multirow{7}{*}[-9pt]{\hspace{18pt}\rotatebox{90}{\textbf{Qwen3-VL}~\cite{bai2025qwen3}}}
            & Baseline~\cite{bai2025qwen3} &  \underline{195.00} {\tiny ($+0.00$)} & 175.00 {\tiny ($+0.00$)} & \underline{168.33} {\tiny ($+0.00$)\phantom{0}} & 193.33 {\tiny ($+0.00$)\phantom{0}} &  731.66 {\tiny ($+0.00$)\phantom{0}} \\
            & Qwen3-VL agent~\cite{bai2025qwen3} &  \textbf{200.00} {\tiny ($+5.00$)} & 181.67 {\tiny ($+6.67$)} & \underline{168.33} {\tiny ($+0.00$)\phantom{0}} & 193.33 {\tiny ($+0.00$)\phantom{0}} &  \underline{743.33} {\tiny ($+11.67$)\phantom{0}} \\ \cmidrule(lr){2-7}
             & VCD~\cite{damonlpsg2023vcd}  & \underline{195.00} {\tiny ($+0.00$)} & 180.00 {\tiny ($+5.00$)\phantom{0}} & \underline{168.33} {\tiny ($+6.73$)\phantom{0}} & 193.33 {\tiny ($+0.00$)} &  736.66 {\tiny ($+5.00$)} \\
             & \RebuttalRevision{Woodpecker}~\cite{yin2024woodpecker} & \RebuttalRevision{\underline{195.00} {\tiny ($+0.00$)}} & \RebuttalRevision{173.33 {\tiny ($+1.67$)\phantom{0}}} & \RebuttalRevision{\underline{168.33} {\tiny ($+0.00$)\phantom{0}}} & \RebuttalRevision{\underline{195.00} {\tiny ($+1.67$)}} &  \RebuttalRevision{731.66 {\tiny ($+10.00$)}} \\
             & RITUAL~\cite{woo2024ritual}  & \underline{195.00} {\tiny ($+0.00$)} & 180.00 {\tiny ($+5.00$)\phantom{0}}  & \underline{168.33} {\tiny ($+0.00$)} & 193.33 {\tiny ($+0.00$)\phantom{0}} & 736.66 {\tiny ($+5.00$)} \\
             & OPERA ~\cite{huang2024opera} & \underline{195.00} {\tiny ($+0.00$)} & 180.00 {\tiny ($+5.00$)\phantom{0}} & \underline{168.33} {\tiny ($+0.00$)\phantom{0}} & \textbf{200.00} {\tiny ($+6.67$)\phantom{0}}  &  \underline{743.33} {\tiny ($+11.67$)} \\
             & DeGF~\cite{zhang2025self}  & \underline{195.00} {\tiny ($+0.00$)} & \underline{181.67} {\tiny ($+6.67$)\phantom{0}} & 166.67 {\tiny ($+1.66$)\phantom{0}} & 188.33 {\tiny ($+5.00$)} &  732.67 {\tiny ($+1.01$)} \\
            & {\textbf{\name}}$_{\mathrm{(ours)}}$ & \cc \textbf{200.00} {\tiny ($+5.00$)} & \cc \textbf{186.67} {\tiny ($+11.67$)\phantom{0}}  & \cc \textbf{180.00} {\tiny ($+11.67$)\phantom{0}} & \cc 193.33 {\tiny ($+0.00$)\phantom{0}} & \cc \textbf{760.00} {\tiny ($+28.34$)\phantom{0}} \\
             \midrule
            \multirow{7}{*}[-2pt]{\hspace{18pt}\rotatebox{90}{\textbf{InternVL3.5}~\cite{wang2025internvl3}}}
            & Baseline~\cite{wang2025internvl3} &  \textbf{200.00} {\tiny ($+0.00$)} & \underline{175.00} {\tiny ($+0.00$)} & \underline{175.00} {\tiny ($+0.00$)\phantom{0}} & \underline{193.33} {\tiny ($+0.00$)\phantom{0}} &  \underline{743.33} {\tiny ($+0.00$)\phantom{0}} \\
             & VCD~\cite{damonlpsg2023vcd}  & \textbf{200.00} {\tiny ($+0.00$)} & \underline{175.00} {\tiny ($+0.00$)\phantom{0}} & \underline{175.00} {\tiny ($+0.00$)\phantom{0}} & \underline{193.33} {\tiny ($+0.00$)} &  736.66 {\tiny ($-6.67$)} \\
             & \RebuttalRevision{Woodpecker}~\cite{yin2024woodpecker} & \RebuttalRevision{\textbf{200.00} {\tiny ($+0.00$)}} & \RebuttalRevision{166.67 {\tiny ($-8.33$)\phantom{0}}} & \RebuttalRevision{161.67 {\tiny ($-13.33$)\phantom{0}}} & \RebuttalRevision{186.67 {\tiny ($-6.66$)}} &  \RebuttalRevision{715.01 {\tiny ($-28.32$)}} \\
             & RITUAL~\cite{woo2024ritual}  & \underline{195.00} {\tiny ($-5.00$)} & \underline{175.00} {\tiny ($+0.00$)\phantom{0}}  & \underline{175.00} {\tiny ($+0.00$)} & \underline{193.33} {\tiny ($+0.00$)\phantom{0}} & 738.33 {\tiny ($-5.00$)} \\
             & OPERA ~\cite{huang2024opera} & \underline{195.00} {\tiny ($-5.00$)} & 173.33 {\tiny ($-1.67$)\phantom{0}} & \underline{175.00} {\tiny ($+0.00$)\phantom{0}} & \textbf{195.00} {\tiny ($+1.67$)\phantom{0}}  &  738.33 {\tiny ($-5.00$)} \\
             & DeGF~\cite{zhang2025self}  & \underline{195.00} {\tiny ($-5.00$)} & \underline{175.00} {\tiny ($+0.00$)\phantom{0}} & 168.33 {\tiny ($-6.67$)\phantom{0}} & 188.33 {\tiny ($-5.00$)} &  726.66 {\tiny ($-16.67$)} \\
            & {\textbf{\name}}$_{\mathrm{(ours)}}$ & \cc \textbf{200.00} {\tiny ($+0.00$)} & \cc \textbf{186.67} {\tiny ($+11.67$)\phantom{0}}  & \cc \textbf{181.67} {\tiny ($+6.67$)\phantom{0}} & \cc \textbf{195.00} {\tiny ($+1.67$)\phantom{0}} & \cc \textbf{763.34} {\tiny ($+20.01$)\phantom{0}} \\
            \bottomrule
        \end{tabular}
        }
        \end{small}
        \end{center}
        
\end{table*}

Tab.~\ref{tab:MME} reports MME-Hallucination results decomposed into object-level (Existence, Count) and attribute-level (Position, Color) subsets.
Overall, \textbf{\name} achieves the best \textbf{MME Score} under Qwen3-VL, reaching \textbf{760.00} ($+ 28.34$), outperforming baselines such as OPERA (\underline{743.33} $+ 11.67$) and VCD/RITUAL (736.66 $+ 5.00$). 
These results suggest that \name’s evidence-driven, claim-level verification and conservative update rule particularly strengthen object existence/count and spatial reasoning, yielding more balanced improvements across hallucination-sensitive attributes. 
We also observe slightly larger variance for \name on some subsets, which is expected when grounding and verification are performed with stochastic components; nevertheless, the mean improvement remains consistent and substantial over all compared methods. Additional results are provided in the supplementary material.

\begin{table*}[t]
\centering
\caption{\textbf{Efficiency comparison.} We report the average inference latency per instance, peak GPU memory, and POPE accuracy on MS-COCO. Experiments are conducted on a single NVIDIA A100 GPU. Lower latency/memory is better, while higher POPE accuracy is better.}
\label{tab:efficiency}
\small
\setlength{\tabcolsep}{10pt}
\resizebox{\linewidth}{!}{
\begin{tabular}{lcccc}
\toprule
\textbf{Method} & \textbf{Average Latency} $\downarrow$ & \textbf{GPU Memory} $\downarrow$ & \textbf{Checked Case} & \textbf{POPE (MS-COCO)} $\uparrow$ \\
\midrule
Qwen3-VL~\cite{bai2025qwen3} & 0.78 s {\tiny ($\times$1.00)} & 17428 MB {\tiny ($\times$1.00)} & 9000 & 87.37 \\
\name ($1^{\text{st}}$ iteration) & 12.00 s {\tiny ($\times$15.38)} & 21472 MB {\tiny ($\times$1.23)} & 9000 & 89.28 \\
\name ($2^{\text{nd}}$ iteration) & 8.54 s {\tiny ($\times$10.94)} & 21296 MB {\tiny ($\times$1.22)} & 4978 & \textbf{89.45} \\
\name ($3^{\text{rd}}$ iteration) & 7.12 s {\tiny ($\times$9.12)} & 21274 MB {\tiny ($\times$1.22)} & 477 & \underline{89.34} \\
\midrule
\cc \textbf{\name} & \cc 18.75 s {\tiny ($\times$24.03)} & \cc 21472 MB {\tiny ($\times$1.23)} & \cc 9000 & \cc \textbf{89.45} \\
\bottomrule
\end{tabular}
}
\end{table*}

Tab.~\ref{tab:MME} presents the results of \name with the InternVL3.5 backbone~\cite{wang2025internvl3} on MME Hallucination. Notably, \name still achieves further improvement over such a strong backbone, showing that our framework remains effective even when the base model already performs at a high level. This result is particularly meaningful, as reducing hallucinations becomes increasingly difficult as the backbone grows stronger. Moreover, among the methods compared in Tab.~\ref{tab:MME}, \name is the only one that consistently improves InternVL3.5, further highlighting the advantage of our grounded evidence-based refinement framework.

\paragraph{Qualitative Analysis}
In Fig.~\ref{fig_exp}, we present results of hallucination mitigation across existence, count, color, and position. The initial LVLM answers are inconsistent with the image content, but are corrected by \name through external evidence from the grounding agent and multi-round self-refinement. The visual evidence provides explicit support for target localization and attribute verification, which helps recover missed objects, disambiguate object counts, reject unsupported color predictions, and correct erroneous spatial relations. Based on these grounded cues, the iterative refinement procedure progressively updates the answer through claim-level verification, yielding more reliable corrections than direct one-step revision. These results demonstrate that explicit grounding and conservative iterative refinement work together to effectively reduce hallucinations across diverse perception scenarios.

\subsection{Efficiency}

Tab.~\ref{tab:efficiency} reports the efficiency of Qwen3-VL~\cite{bai2025qwen3} and \name in terms of average inference latency, peak GPU memory, checked cases, and POPE accuracy on MS-COCO~\cite{lin2014microsoft}. As expected, \name incurs substantially higher end-to-end latency than a single-pass LVLM inference, due to the additional overhead introduced by external grounding, claim-level verification, and iterative self-refinement. 
Importantly, the per-round statistics are computed over different numbers of checked cases because of early stopping: the first iteration is applied to all 9000 cases, whereas only 4978 and 477 cases proceed to the second and third iterations, respectively. Consequently, later iterations are evaluated on progressively smaller subsets. Since many easy cases are resolved early, the remaining instances typically involve fewer unresolved claims and require less additional evidence, resulting in lower average latency and slightly reduced memory usage in subsequent iterations. This also explains why the best POPE accuracy is already achieved after the second iteration, while the third iteration only yields limited changes on a much smaller subset. Based on this trade-off, we set the maximum number of refinement iterations to 3, which provides a practical balance between performance and efficiency.

\subsection{Human Study}

To complement the automatic benchmark results, we conduct a human study to evaluate whether the responses produced by \name are better aligned with human preference under evidence-grounded judgment. 
To complement automatic benchmarks, we conduct a human preference study to assess whether \name produces responses that are more reliable and more interpretable.

\begin{figure}[h!]
    \centering
    \includegraphics[width=1\linewidth]{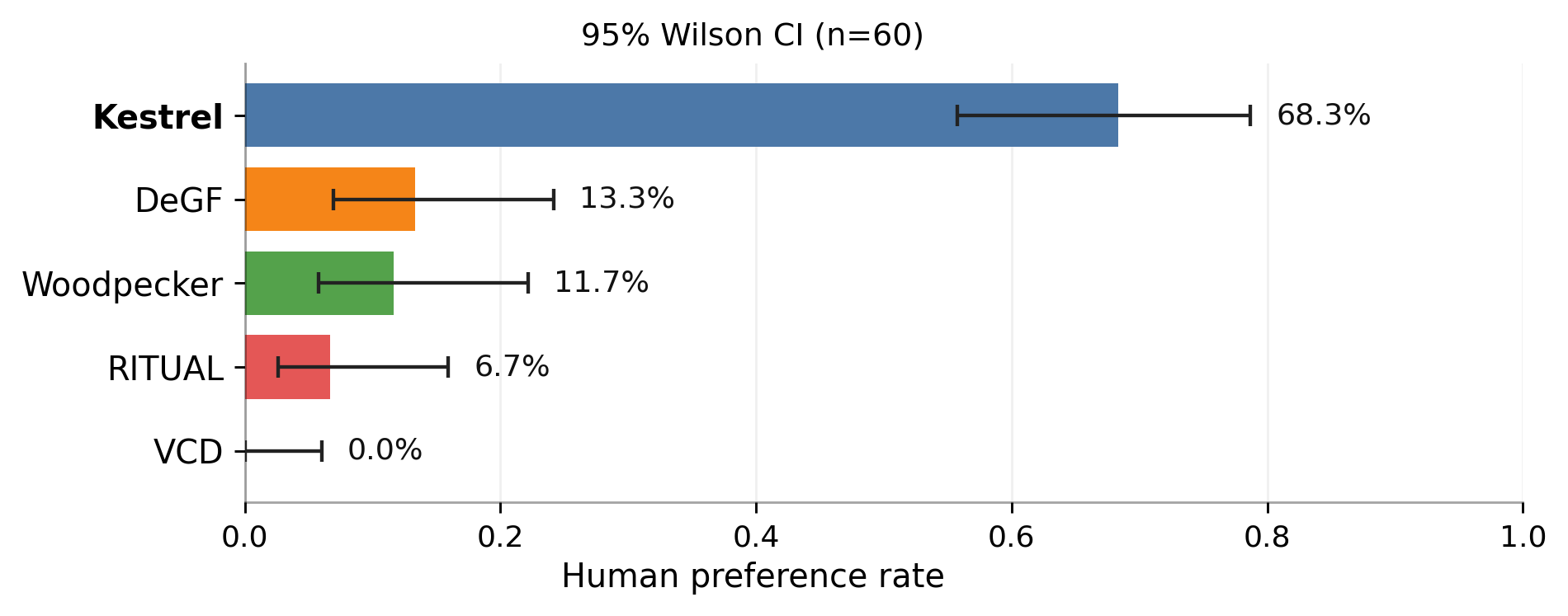}
     \caption{
     \textbf{Human preference comparison across methods.} Bars show the fraction of trials in which each method is preferred (n=60). Error bars denote 95$\%$ Wilson confidence intervals.
     }
     \label{fig_human}
   \end{figure}

\paragraph{Setup.} 
We sample 60 evaluation cases and compare five training-free mitigation methods: \name, DeGF, Woodpecker, RITUAL, and VCD, which are also included in our main experiments. For each case, annotators are presented with the candidate outputs in randomized order and asked to choose the response they prefer based on grounding quality, factual consistency with the image, and overall answer reliability. We report the human preference rate of each method.

\paragraph{Results.} 
As shown in Fig.~\ref{fig_human}, \name is preferred in 41/60 cases (68.3$\%$), substantially outperforming DeGF (13.3$\%$), Woodpecker (11.7$\%$), RITUAL (6.7$\%$), and VCD (0.0$\%$). This result indicates that \name’s outputs are more consistent with human judgments of groundedness and answer quality. We attribute this advantage to \name’s explicit evidence-backed verification process. Instead of relying only on decoding-time intervention, \name externalizes grounding into structured visual and textual evidence, verifies question-aligned claims against cited evidence, and updates the answer only when the evidence is sufficiently reliable. 

\subsection{Ablation Study}
\label{sec:ablation}

\begin{table*}[t]
\centering
\caption{\textbf{Ablation study of core design components.}
We evaluate different settings by progressively enabling the key components in \name. Higher scores indicate better performance.}
\label{tab:ablation_core}
\small
\setlength{\tabcolsep}{10pt}
\resizebox{\linewidth}{!}{
\begin{tabular}{c cccccc c}
\toprule
\multirow{2}{*}{\textbf{Setting}}
& \textbf{Grounding} 
& \textbf{Structured Textual} 
& \textbf{Claim-level} 
& \textbf{Evidence-gated} 
& \textbf{Self-} 
& \textbf{History-} 
& \textbf{POPE} \\
& \textbf{Agent}
& \textbf{Evidence}
& \textbf{Verification}
& \textbf{Update}
& \textbf{Refinement}
& \textbf{Aware} 
& \textbf{MS-COCO} \\
\midrule
baseline & \xmark & \xmark & \xmark & \xmark & \xmark & \xmark & 87.37 \\
s1       & \xmark & \xmark & \xmark & \xmark & \cmark & \xmark & 87.44 \\
s2       & \cmark & \xmark & \xmark & \xmark & \xmark & \xmark & 87.50 \\
s3       & \xmark & \xmark & \cmark & \cmark & \cmark & \cmark & 87.58 \\
s4       & \cmark & \xmark & \cmark & \cmark & \cmark & \cmark & 88.61 \\
s5       & \cmark & \cmark & \xmark & \cmark & \cmark & \cmark & 89.05 \\
s6       & \cmark & \cmark & \cmark & \xmark & \cmark & \cmark & 89.30 \\
s7       & \cmark & \cmark & \cmark & \cmark & \xmark & \xmark & 89.16 \\
s8       & \cmark & \cmark & \cmark & \cmark & \cmark & \xmark & 89.32 \\
\cc s9   & \cc \cmark & \cc \cmark & \cc \cmark & \cc \cmark & \cc \cmark & \cc \cmark & \cc \textbf{89.45} \\
\bottomrule
\end{tabular}
}
\end{table*}

\begin{figure*}[t]
\begin{minipage}[t]{0.49\linewidth}
\makeatletter\def\@captype{table}
\setlength\tabcolsep{11.5pt}
\caption{\textbf{Ablation of visual evidence types.} Contribution of different visual evidence forms, including segmentation, bounding boxes, and crop-zoom views. Higher scores indicate better performance.}
    \label{tab:vis}
    \resizebox{\textwidth}{!}{
    \begin{tabular}{lcccc}
        \toprule
        visual & segment & box & crop-zoom & MME Score \\
        \midrule
        baseline & -- & -- & -- & 731.66 \\
        s1 & \xmark & \cmark & \cmark & 738.33 \\
        s2 & \cmark & \xmark & \cmark & 743.33 \\
        s3 & \cmark & \cmark & \xmark & 740.00 \\
        \cc s4 & \cc\cmark & \cc\cmark & \cc\cmark & \cc760.00 \\
        \bottomrule
    \end{tabular}
    }
\end{minipage}
\begin{minipage}[t]{0.49\linewidth}
\makeatletter\def\@captype{table}
\setlength\tabcolsep{10pt}
\caption{\textbf{Ablation of structured textual evidence.} We study the contribution of different textual evidence types used for claim verification. Higher scores indicate better performance.}
    \label{tab:tex}
    \resizebox{\textwidth}{!}{
    \begin{tabular}{lccccc}
        \toprule
        textual & exist & count & color & position & MME Score \\
        \midrule
        s1 & \xmark & \cmark & \cmark & \cmark & 755.00 \\
        s2 & \cmark & \xmark & \cmark & \cmark & 751.67 \\
        s3 & \cmark & \cmark & \xmark & \cmark & 753.33 \\
        s4 & \cmark & \cmark & \cmark & \xmark & 748.33 \\
         \cc s5 & \cc\cmark & \cc\cmark & \cc\cmark & \cc\cmark & \cc760.00 \\
        \bottomrule
    \end{tabular}
    }
\end{minipage}
\vspace{-8pt}
\end{figure*}

\paragraph{Designs.} 
Tab.~\ref{tab:ablation_core} reports the ablation results of \name on POPE-COCO. Naive self-refinement and the grounding agent alone bring only marginal improvements over the baseline, indicating that neither iterative revision nor external tool access is sufficient by itself. Introducing verification-guided refinement further improves performance, while incorporating the grounding agent yields a larger gain, highlighting the importance of explicit grounded evidence. Adding structured textual evidence and claim-level verification leads to further improvements, showing that normalized evidence construction and fine-grained verification are both beneficial. We also observe that removing evidence-gated update degrades performance, confirming the role of conservative update control in preventing risky corrections. Finally, self-refinement and stateful refinement provide additional gains on top of the full evidence-verification pipeline, and the full \name achieves the best result overall.

\paragraph{Evidence.}

The evidence ablations (Tab.~\ref{tab:vis} and Tab.~\ref{tab:tex}) further highlight the complementary roles of both visual and textual evidence in \name. For visual evidence, removing any single component consistently degrades performance relative to the full setting, showing that segmentation overlays (segment), bounding boxes (box), and crop-zoom views (crop-zoom) all contribute to the final result. Among them, removing segmentation or crop-zoom causes a larger drop than removing boxes, suggesting that precise region localization and enlarged local inspection are particularly important for reliable claim verification. For structured textual evidence, ablating any evidence type also leads to performance degradation, confirming that existence, count, color, and position evidence all provide useful support for refinement. Notably, removing position or count evidence causes a relatively larger drop, indicating that these attributes benefit more from explicit structured evidence. Overall, the best performance is achieved when all visual and textual evidence types are used together, demonstrating that the different evidence forms are complementary rather than redundant.

\section{Conclusion}

In this paper, we presented \name, a training-free framework for LVLM hallucination mitigation that integrates explicit visual grounding with structured evidence-driven self-refinement. By converting external tool outputs into citable visual and textual evidence, \name enables claim-level verification and conservative evidence-gated answer updates, providing both improved reliability and transparent verification traces. Extensive experiments on POPE and MME-Hallucination show that \name consistently improves over strong training-free baselines across multiple backbones, with particularly clear gains under more challenging hallucination settings. Further ablations validate the contributions of each design. Overall, \name demonstrates that coupling explicit grounding with interpretable iterative verification offers an effective and stable path toward more trustworthy LVLMs.

{
    \small
    \bibliographystyle{ieeenat_fullname}
    \bibliography{main}

@inproceedings{radford2021learning,
  title={Learning transferable visual models from natural language supervision},
  author={Radford, Alec and Kim, Jong Wook and Hallacy, Chris and Ramesh, Aditya and Goh, Gabriel and Agarwal, Sandhini and Sastry, Girish and Askell, Amanda and Mishkin, Pamela and Clark, Jack and others},
  booktitle={International conference on machine learning},
  pages={8748--8763},
  year={2021},
  organization={PmLR}
}

@inproceedings{zhai2023sigmoid,
  title={Sigmoid loss for language image pre-training},
  author={Zhai, Xiaohua and Mustafa, Basil and Kolesnikov, Alexander and Beyer, Lucas},
  booktitle={Proceedings of the IEEE/CVF international conference on computer vision},
  pages={11975--11986},
  year={2023}
}

@article{alayrac2022flamingo,
  title={Flamingo: a visual language model for few-shot learning},
  author={Alayrac, Jean-Baptiste and Donahue, Jeff and Luc, Pauline and Miech, Antoine and Barr, Iain and Hasson, Yana and Lenc, Karel and Mensch, Arthur and Millican, Katherine and Reynolds, Malcolm and others},
  journal={Advances in neural information processing systems},
  volume={35},
  pages={23716--23736},
  year={2022}
}

@article{liu2023visual,
  title={Visual instruction tuning},
  author={Liu, Haotian and Li, Chunyuan and Wu, Qingyang and Lee, Yong Jae},
  journal={Advances in neural information processing systems},
  volume={36},
  pages={34892--34916},
  year={2023}
}

@article{dai2023instructblip,
  title={Instructblip: Towards general-purpose vision-language models with instruction tuning},
  author={Dai, Wenliang and Li, Junnan and Li, Dongxu and Tiong, Anthony and Zhao, Junqi and Wang, Weisheng and Li, Boyang and Fung, Pascale N and Hoi, Steven},
  journal={Advances in neural information processing systems},
  volume={36},
  pages={49250--49267},
  year={2023}
}

@article{bai2025qwen3,
  title={Qwen3-vl technical report},
  author={Bai, Shuai and Cai, Yuxuan and Chen, Ruizhe and Chen, Keqin and Chen, Xionghui and Cheng, Zesen and Deng, Lianghao and Ding, Wei and Gao, Chang and Ge, Chunjiang and others},
  journal={arXiv preprint arXiv:2511.21631},
  year={2025}
}

@article{wang2025internvl3,
  title={Internvl3. 5: Advancing open-source multimodal models in versatility, reasoning, and efficiency},
  author={Wang, Weiyun and Gao, Zhangwei and Gu, Lixin and Pu, Hengjun and Cui, Long and Wei, Xingguang and Liu, Zhaoyang and Jing, Linglin and Ye, Shenglong and Shao, Jie and others},
  journal={arXiv preprint arXiv:2508.18265},
  year={2025}
}

@article{grattafiori2024llama,
  title={The llama 3 herd of models},
  author={Grattafiori, Aaron and Dubey, Abhimanyu and Jauhri, Abhinav and Pandey, Abhinav and Kadian, Abhishek and Al-Dahle, Ahmad and Letman, Aiesha and Mathur, Akhil and Schelten, Alan and Vaughan, Alex and others},
  journal={arXiv preprint arXiv:2407.21783},
  year={2024}
}

@article{li2023evaluating,
  title={Evaluating object hallucination in large vision-language models},
  author={Li, Yifan and Du, Yifan and Zhou, Kun and Wang, Jinpeng and Zhao, Wayne Xin and Wen, Ji-Rong},
  journal={arXiv preprint arXiv:2305.10355},
  year={2023}
}

@article{wang2023amber,
  title={Amber: An llm-free multi-dimensional benchmark for mllms hallucination evaluation},
  author={Wang, Junyang and Wang, Yuhang and Xu, Guohai and Zhang, Jing and Gu, Yukai and Jia, Haitao and Wang, Jiaqi and Xu, Haiyang and Yan, Ming and Zhang, Ji and others},
  journal={arXiv preprint arXiv:2311.07397},
  year={2023}
}

@inproceedings{sun2024aligning,
  title={Aligning large multimodal models with factually augmented rlhf},
  author={Sun, Zhiqing and Shen, Sheng and Cao, Shengcao and Liu, Haotian and Li, Chunyuan and Shen, Yikang and Gan, Chuang and Gui, Liangyan and Wang, Yu-Xiong and Yang, Yiming and others},
  booktitle={Findings of the Association for Computational Linguistics: ACL 2024},
  pages={13088--13110},
  year={2024}
}

@article{damonlpsg2023vcd,
  author  = {Leng, Sicong and Zhang, Hang and Chen, Guanzheng and Li, Xin and Lu, Shijian and Miao, Chunyan and Bing, Lidong},
  title   = {Mitigating Object Hallucinations in Large Vision-Language Models through Visual Contrastive Decoding},
  journal = {arXiv preprint arXiv:2311.16922},
  year    = {2023},
  url     = {https://arxiv.org/abs/2311.16922}
}

@inproceedings{huang2024opera,
  title={Opera: Alleviating hallucination in multi-modal large language models via over-trust penalty and retrospection-allocation},
  author={Huang, Qidong and Dong, Xiaoyi and Zhang, Pan and Wang, Bin and He, Conghui and Wang, Jiaqi and Lin, Dahua and Zhang, Weiming and Yu, Nenghai},
  booktitle={Proceedings of the IEEE/CVF Conference on Computer Vision and Pattern Recognition},
  pages={13418--13427},
  year={2024}
}

@article{zhang2025self,
  title={Self-correcting decoding with generative feedback for mitigating hallucinations in large vision-language models},
  author={Zhang, Ce and Wan, Zifu and Kan, Zhehan and Ma, Martin Q and Stepputtis, Simon and Ramanan, Deva and Salakhutdinov, Russ and Morency, Louis-Philippe and Sycara, Katia and Xie, Yaqi},
  journal={arXiv preprint arXiv:2502.06130},
  year={2025}
}

@article{huang2025shield,
  title={SHIELD: Suppressing Hallucinations In LVLM Encoders via Bias and Vulnerability Defense},
  author={Huang, Yiyang and Shi, Liang and Zhang, Yitian and Xu, Yi and Fu, Yun},
  journal={arXiv preprint arXiv:2510.16596},
  year={2025}
}

@article{yin2024woodpecker,
  title={Woodpecker: Hallucination correction for multimodal large language models},
  author={Yin, Shukang and Fu, Chaoyou and Zhao, Sirui and Xu, Tong and Wang, Hao and Sui, Dianbo and Shen, Yunhang and Li, Ke and Sun, Xing and Chen, Enhong},
  journal={Science China Information Sciences},
  volume={67},
  number={12},
  pages={220105},
  year={2024},
  publisher={Springer}
}

@article{woo2024ritual,
  title   = {{RITUAL}: Random Image Transformations as a Universal Anti-hallucination Lever in Large Vision Language Models},
  author  = {Woo, Sangmin and Jang, Jaehyuk and Kim, Donguk and Choi, Yubin and Kim, Changick},
  journal = {arXiv preprint arXiv:2405.17821},
  year    = {2024},
  url     = {https://arxiv.org/abs/2405.17821}
}

@misc{carion2025sam3segmentconcepts,
      title={SAM 3: Segment Anything with Concepts},
      author={Nicolas Carion and Laura Gustafson and Yuan-Ting Hu and Shoubhik Debnath and Ronghang Hu and Didac Suris and Chaitanya Ryali and Kalyan Vasudev Alwala and Haitham Khedr and Andrew Huang and Jie Lei and Tengyu Ma and Baishan Guo and Arpit Kalla and Markus Marks and Joseph Greer and Meng Wang and Peize Sun and Roman Rädle and Triantafyllos Afouras and Effrosyni Mavroudi and Katherine Xu and Tsung-Han Wu and Yu Zhou and Liliane Momeni and Rishi Hazra and Shuangrui Ding and Sagar Vaze and Francois Porcher and Feng Li and Siyuan Li and Aishwarya Kamath and Ho Kei Cheng and Piotr Dollár and Nikhila Ravi and Kate Saenko and Pengchuan Zhang and Christoph Feichtenhofer},
      year={2025},
      eprint={2511.16719},
      archivePrefix={arXiv},
      primaryClass={cs.CV},
      url={https://arxiv.org/abs/2511.16719},
}

@article{fu2023mme,
  title={Mme: A comprehensive evaluation benchmark for multimodal large language models},
  author={Fu, Chaoyou and Chen, Peixian and Shen, Yunhang and Qin, Yulei and Zhang, Mengdan and Lin, Xu and Yang, Jinrui and Zheng, Xiawu and Li, Ke and Sun, Xing and others},
  journal={arXiv preprint arXiv:2306.13394},
  year={2023}
}

@inproceedings{lin2014microsoft,
  title={Microsoft coco: Common objects in context},
  author={Lin, Tsung-Yi and Maire, Michael and Belongie, Serge and Hays, James and Perona, Pietro and Ramanan, Deva and Doll{\'a}r, Piotr and Zitnick, C Lawrence},
  booktitle={European conference on computer vision},
  pages={740--755},
  year={2014},
  organization={Springer}
}

@inproceedings{schwenk2022okvqa,
  title={A-okvqa: A benchmark for visual question answering using world knowledge},
  author={Schwenk, Dustin and Khandelwal, Apoorv and Clark, Christopher and Marino, Kenneth and Mottaghi, Roozbeh},
  booktitle={European conference on computer vision},
  pages={146--162},
  year={2022},
  organization={Springer}
}

@inproceedings{hudson2019gqa,
  title={Gqa: A new dataset for real-world visual reasoning and compositional question answering},
  author={Hudson, Drew A and Manning, Christopher D},
  booktitle={Proceedings of the IEEE/CVF conference on computer vision and pattern recognition},
  pages={6700--6709},
  year={2019}
}

@inproceedings{zhang2024reflective,
  title={Reflective instruction tuning: Mitigating hallucinations in large vision-language models},
  author={Zhang, Jinrui and Wang, Teng and Zhang, Haigang and Lu, Ping and Zheng, Feng},
  booktitle={European Conference on Computer Vision},
  pages={196--213},
  year={2024},
  organization={Springer}
}

@article{chen2023mitigating,
  title={Mitigating hallucination in visual language models with visual supervision},
  author={Chen, Zhiyang and Zhu, Yousong and Zhan, Yufei and Li, Zhaowen and Zhao, Chaoyang and Wang, Jinqiao and Tang, Ming},
  journal={arXiv preprint arXiv:2311.16479},
  year={2023}
}

@inproceedings{jiang2024hallucination,
  title={Hallucination augmented contrastive learning for multimodal large language model},
  author={Jiang, Chaoya and Xu, Haiyang and Dong, Mengfan and Chen, Jiaxing and Ye, Wei and Yan, Ming and Ye, Qinghao and Zhang, Ji and Huang, Fei and Zhang, Shikun},
  booktitle={Proceedings of the IEEE/CVF Conference on Computer Vision and Pattern Recognition},
  pages={27036--27046},
  year={2024}
}

@article{lyu2024alleviating,
  title={Alleviating hallucinations in large vision-language models through hallucination-induced optimization},
  author={Lyu, Xinyu and Chen, Beitao and Gao, Lianli and Shen, Hengtao and Song, Jingkuan},
  journal={Advances in Neural Information Processing Systems},
  volume={37},
  pages={122811--122832},
  year={2024}
}

@inproceedings{guan2024hallusionbench,
  title={Hallusionbench: an advanced diagnostic suite for entangled language hallucination and visual illusion in large vision-language models},
  author={Guan, Tianrui and Liu, Fuxiao and Wu, Xiyang and Xian, Ruiqi and Li, Zongxia and Liu, Xiaoyu and Wang, Xijun and Chen, Lichang and Huang, Furong and Yacoob, Yaser and others},
  booktitle={Proceedings of the IEEE/CVF conference on computer vision and pattern recognition},
  pages={14375--14385},
  year={2024}
}

@inproceedings{li2023blip,
  title={Blip-2: Bootstrapping language-image pre-training with frozen image encoders and large language models},
  author={Li, Junnan and Li, Dongxu and Savarese, Silvio and Hoi, Steven},
  booktitle={International conference on machine learning},
  pages={19730--19742},
  year={2023},
  organization={PMLR}
}

@inproceedings{liu2024improved,
  title={Improved baselines with visual instruction tuning},
  author={Liu, Haotian and Li, Chunyuan and Li, Yuheng and Lee, Yong Jae},
  booktitle={Proceedings of the IEEE/CVF conference on computer vision and pattern recognition},
  pages={26296--26306},
  year={2024}
}

@article{awadalla2023openflamingo,
  title={Openflamingo: An open-source framework for training large autoregressive vision-language models},
  author={Awadalla, Anas and Gao, Irena and Gardner, Josh and Hessel, Jack and Hanafy, Yusuf and Zhu, Wanrong and Marathe, Kalyani and Bitton, Yonatan and Gadre, Samir and Sagawa, Shiori and others},
  journal={arXiv preprint arXiv:2308.01390},
  year={2023}
}

@article{wang2024cogvlm,
  title={Cogvlm: Visual expert for pretrained language models},
  author={Wang, Weihan and Lv, Qingsong and Yu, Wenmeng and Hong, Wenyi and Qi, Ji and Wang, Yan and Ji, Junhui and Yang, Zhuoyi and Zhao, Lei and XiXuan, Song and others},
  journal={Advances in Neural Information Processing Systems},
  volume={37},
  pages={121475--121499},
  year={2024}
}

@article{peng2023kosmos,
  title={Kosmos-2: Grounding multimodal large language models to the world},
  author={Peng, Zhiliang and Wang, Wenhui and Dong, Li and Hao, Yaru and Huang, Shaohan and Ma, Shuming and Wei, Furu},
  journal={arXiv preprint arXiv:2306.14824},
  year={2023}
}

@inproceedings{kaul2024throne,
  title={Throne: An object-based hallucination benchmark for the free-form generations of large vision-language models},
  author={Kaul, Prannay and Li, Zhizhong and Yang, Hao and Dukler, Yonatan and Swaminathan, Ashwin and Taylor, CJ and Soatto, Stefano},
  booktitle={Proceedings of the IEEE/CVF Conference on Computer Vision and Pattern Recognition},
  pages={27228--27238},
  year={2024}
}

@article{li2023silkie,
  title={Silkie: Preference distillation for large visual language models},
  author={Li, Lei and Xie, Zhihui and Li, Mukai and Chen, Shunian and Wang, Peiyi and Chen, Liang and Yang, Yazheng and Wang, Benyou and Kong, Lingpeng},
  journal={arXiv preprint arXiv:2312.10665},
  year={2023}
}

@misc{chen2024allavaharnessinggpt4vsynthesizeddata,
      title={ALLaVA: Harnessing GPT4V-Synthesized Data for Lite Vision-Language Models}, 
      author={Guiming Hardy Chen and Shunian Chen and Ruifei Zhang and Junying Chen and Xiangbo Wu and Zhiyi Zhang and Zhihong Chen and Jianquan Li and Xiang Wan and Benyou Wang},
      year={2024},
      eprint={2402.11684},
      archivePrefix={arXiv},
      primaryClass={cs.CL},
      url={https://arxiv.org/abs/2402.11684}, 
}

@misc{chen2023sharegpt4vimprovinglargemultimodal,
      title={ShareGPT4V: Improving Large Multi-Modal Models with Better Captions}, 
      author={Lin Chen and Jinsong Li and Xiaoyi Dong and Pan Zhang and Conghui He and Jiaqi Wang and Feng Zhao and Dahua Lin},
      year={2023},
      eprint={2311.12793},
      archivePrefix={arXiv},
      primaryClass={cs.CV},
      url={https://arxiv.org/abs/2311.12793}, 
}
}

\appendix

\section*{Technical Appendices}

\section{Prompt Template}

We provide the prompt templates used in the three core stages of \name: Initialization (Fig.~\ref{fig_hypothesis_prompt}), claim-level verification (Fig.~\ref {fig_verify_prompt}), and self-refinement (Fig.~\ref {fig_refine_prompt}). These prompts follow the main evidence-grounded iterative framework, in which \name first rewrites the original question into a concrete and visually verifiable claim, then verifies the claim against the collected evidence, and finally updates the answer conservatively through refinement. To improve reproducibility, all prompts adopt constrained JSON outputs with explicit field definitions. The initialization prompt produces a question-aligned claim for grounding, the verification prompt returns evidence-based verdicts and confidence scores, and the self-refinement prompt updates the answer while proposing a new claim for the next round.

 \begin{figure*}[t]
  \centering
  \begin{minipage}{1\textwidth}
  \begin{tcolorbox}[
      title={Prompt template for Initialization},
      fontupper=\small,
      fonttitle=\small
  ]
  \raggedright

  \textcolor{blue}{messages}=[
  \{"role":"user","content":[
  \{"type":"image","image":\textcolor{blue}{sample["image"]}\},
  \{"type":"text","text":
  You are given an image and a Yes/No question.
  Determine the answer and output one verifiable claim.

  Task requirements:
  - answer must be exactly "Yes" or "No".
  - output exactly one claim in verifiable\_claims.
  - claim fields: id, type, text, targets.
  - type must be exactly "\textcolor{blue}{sample["expected\_claim\_type"]}".
  - text must be concrete and visually checkable.

  Target rules:
  - if type = position: targets contain 1--2 short object phrases;
    for two-object relation, use [subject, anchor].
  - otherwise: targets contain exactly one object noun from question;
    do not include color, number, or position words.

  Return JSON only:
  \{
    "answer":"Yes|No",
    "verifiable\_claims":[
      \{"id":"c1",
        "type":"\textcolor{blue}{sample["expected\_claim\_type"]}",
        "text":"...",
        "targets":\textcolor{blue}{sample["example\_targets"]}\}
    ]
  \}

  Question: "\textcolor{blue}{sample["question"]}"
  PrevSummary (optional): "\textcolor{blue}{sample["prev\_summary"]}"
  <image>
  "\}
  ]\}]

  \vspace{0.8em}
  \hrule
  \vspace{0.8em}

  \textbf{Yes-Guard Template (optional, only when initial answer = Yes):}

  \textcolor{blue}{messages}=[
  \{"role":"user","content":[
  \{"type":"image","image":\textcolor{blue}{sample["image"]}\},
  \{"type":"text","text":
  Check whether the Yes/No question is true in this image.

  Return JSON only:
  \{
    "answer":"yes|no|unclear",
    "confidence":"high|medium|low",
    "reason":"one short sentence"
  \}

  Rules:
  - use image evidence only;
  - yes: clearly true; no: clearly false; otherwise unclear;
  - be conservative when evidence is weak.

  Question: "\textcolor{blue}{sample["question"]}"
  Target hint: "\textcolor{blue}{sample["target\_hint"]}"
  <image>
  "\}
  ]\}]

  \end{tcolorbox}
  \end{minipage}
  \vspace{-0.5em}
  \caption{Prompt template for initialization stage.}
  \vspace{-0.5em}
  \label{fig_hypothesis_prompt}
  \end{figure*}

\begin{figure*}[t]
  \centering
  \begin{minipage}{1\textwidth}
  \begin{tcolorbox}[
      title={Prompt template for claim-level verification},
      fontupper=\small,
      fonttitle=\small
  ]
  \raggedright


  \textcolor{blue}{messages}=[
  \{"role":"user","content":[
  \textcolor{blue}{sample["evidence\_content"]},
  \{"type":"text","text":
  You are a strict verifier.
  Judge each claim using ONLY the provided evidence items.

  How to use evidence:
  - Each item has an EvidenceID and a Type.
  - Use only EvidenceID values that appear in the given evidence.
  - Typical IDs:
    - seg\_overlay: e\_seg\_\{tkey\}
    - count\_text: e\_count\_\{tkey\}
    - count\_compare\_text: e\_countcmp\_\{tkey\}
    - count\_vision\_text: e\_countvis\_\{tkey\}
    - count\_vision\_compare\_text: e\_countviscmp\_\{tkey\}
    - crop\_zoom: e\_crop\_\{tkey\}
    - color\_text: e\_color\_\{tkey\}
    - position\_text: e\_pos\_\{tkey\}
    - position\_relation\_text: e\_posrel\_\{claim\_id\}

  Judging rules:
  - For each claim, choose exactly one status:
    supported | contradicted | insufficient.
  - supported: evidence clearly confirms the claim.
  - contradicted: evidence clearly refutes the claim.
  - insufficient: evidence is missing or ambiguous.
  - Do NOT use common sense.

  Top-level verdict:
  - contradicted: at least one claim is strongly contradicted.
  - supported: all claims are strongly supported.
  - otherwise: insufficient.

  Return JSON only:
  \{
    "verdict":"supported|contradicted|insufficient",
    "checked":[
      \{
        "claim\_id":"...",
        "status":"supported|contradicted|insufficient",
        "confidence":0.0,
        "why":"...",
        "citations":["EvidenceID1","EvidenceID2"]
      \}
    ]
  \}

  Input:
  Question: "\textcolor{blue}{sample["question"]}"
  Claims: \textcolor{blue}{sample["claims\_json"]}
  PrevVerdict (optional): \textcolor{blue}{sample["prev\_verdict\_json"]}
  "\}
  ]\}]

  \end{tcolorbox}
  \end{minipage}
  \vspace{-0.5em}
  \caption{Prompt template for claim-level verification stage.}
  \vspace{-1.6em}
  \label{fig_verify_prompt}
  \end{figure*}

\begin{figure*}[t]
  \centering
  \begin{minipage}{1\textwidth}
  \begin{tcolorbox}[
      title={Prompt template for Self-Refine},
      fontupper=\small,
      fonttitle=\small
  ]
  \raggedright


  \textcolor{blue}{messages}=[
  \{"role":"user","content":[
  \{"type":"image","image":\textcolor{blue}{sample["image"]}\},
  \{"type":"text","text":
  You are a cautious VQA assistant.
  Decide the final answer as a binary choice.

  HARD REQUIREMENT:
  - Answer must be exactly "Yes" or "No" (capitalized).
  - No punctuation, no explanation, no extra tokens.
  - You must output a concrete answer.

  Use all available history:
  - previous rounds: hypothesis / verify / refine;
  - current round: hypothesis and verify details.

  Also output one new\_claim for the next round.

  Claim constraints:
  - RELEVANCE: every new\_claim must directly verify the Question's Yes/No.
  - new\_claim.type must be exactly
    "\textcolor{blue}{sample["expected\_claim\_type"]}".
  - claim.text must mention at least one key
    entity / attribute / relation from the Question.

  Type-specific target rules:
  - if type = position:
    - targets contain 1--2 short object phrases from Question;
    - if two-object relation exists, use [subject, anchor];
    - keep distinguishing modifiers when needed.
  - otherwise:
    - targets contain exactly one object noun from Question;
    - do not include color / number / position / quantifier words.
  - if a claim is not clearly usable to support/refute Question,
    it is invalid and must not be output.

  Return JSON only:
  \{
    "new\_claims":[
      \{"id":"c1",
        "type":"\textcolor{blue}{sample["expected\_claim\_type"]}",
        "text":"...",
        "targets":\textcolor{blue}{sample["example\_targets"]},
        "priority":1\}
    ],
    "Answer":"Yes|No"
  \}

  Question: "\textcolor{blue}{sample["question"]}"
  PreviousAnswer: "\textcolor{blue}{sample["A\_t"]}"
  RoundHistory: \textcolor{blue}{sample["round\_history\_json"]}
  CurrentRoundContext: \textcolor{blue}{sample["current\_round\_context\_json"]}
  \textless image\textgreater
  "\}
  ]\}]

  \end{tcolorbox}
  \end{minipage}
  \vspace{-0.5em}
  \caption{Prompt template for self-refinement stage.}
  \vspace{-0.5em}
  \label{fig_refine_prompt}
  \end{figure*}




\section{More Case Studies}

\begin{figure*}[h!]
    \centering
    \includegraphics[width=1\linewidth]{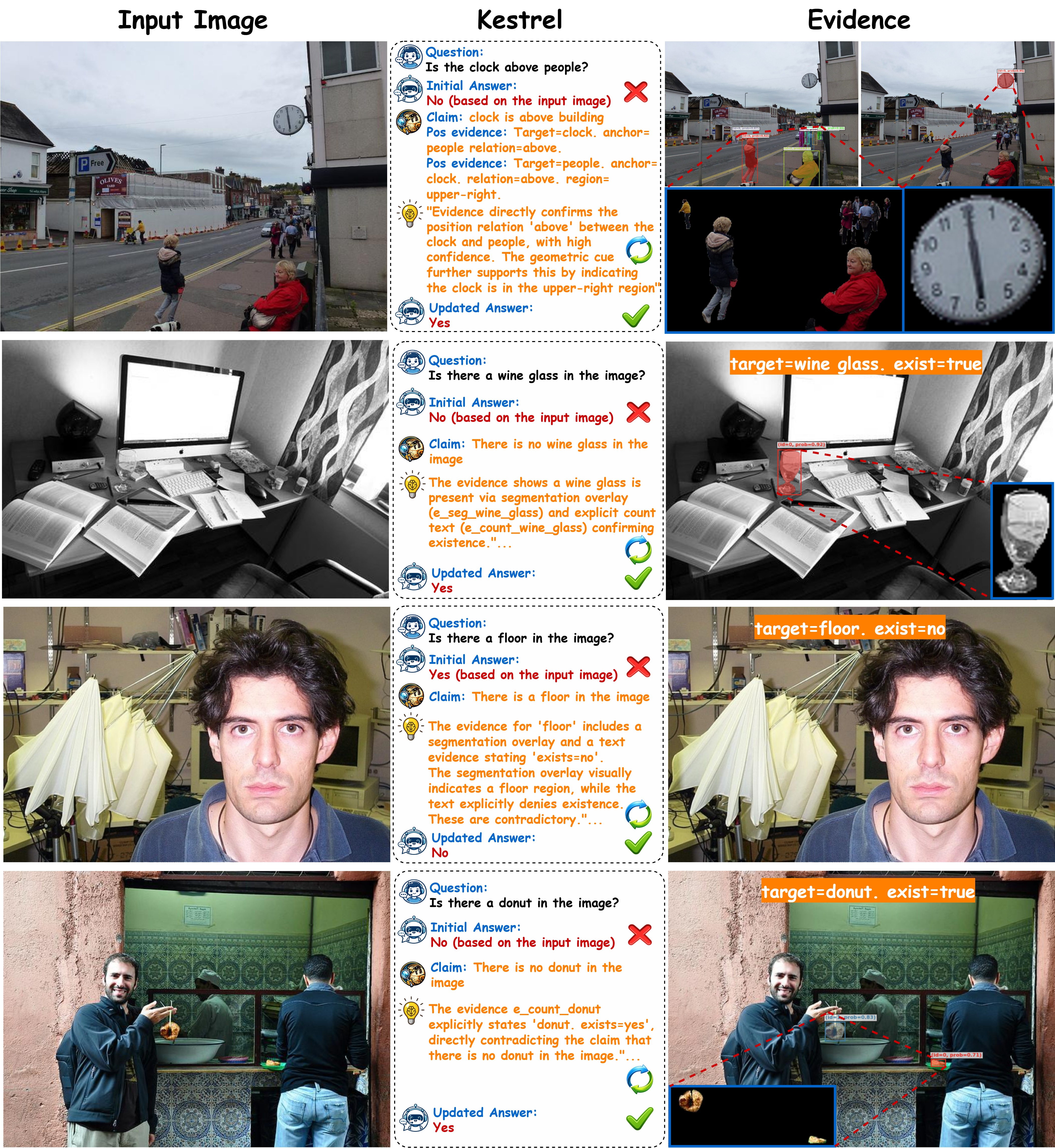}
     \caption{
     \textbf{Qualitative results of \name.} We compare the VQA responses from the regular baseline and our method based on Qwen3-VL. Zoom in for a better view.
     }
     \label{fig_app_q}
   \end{figure*}

\begin{figure*}[h!]
    \centering
    \includegraphics[width=1\linewidth]{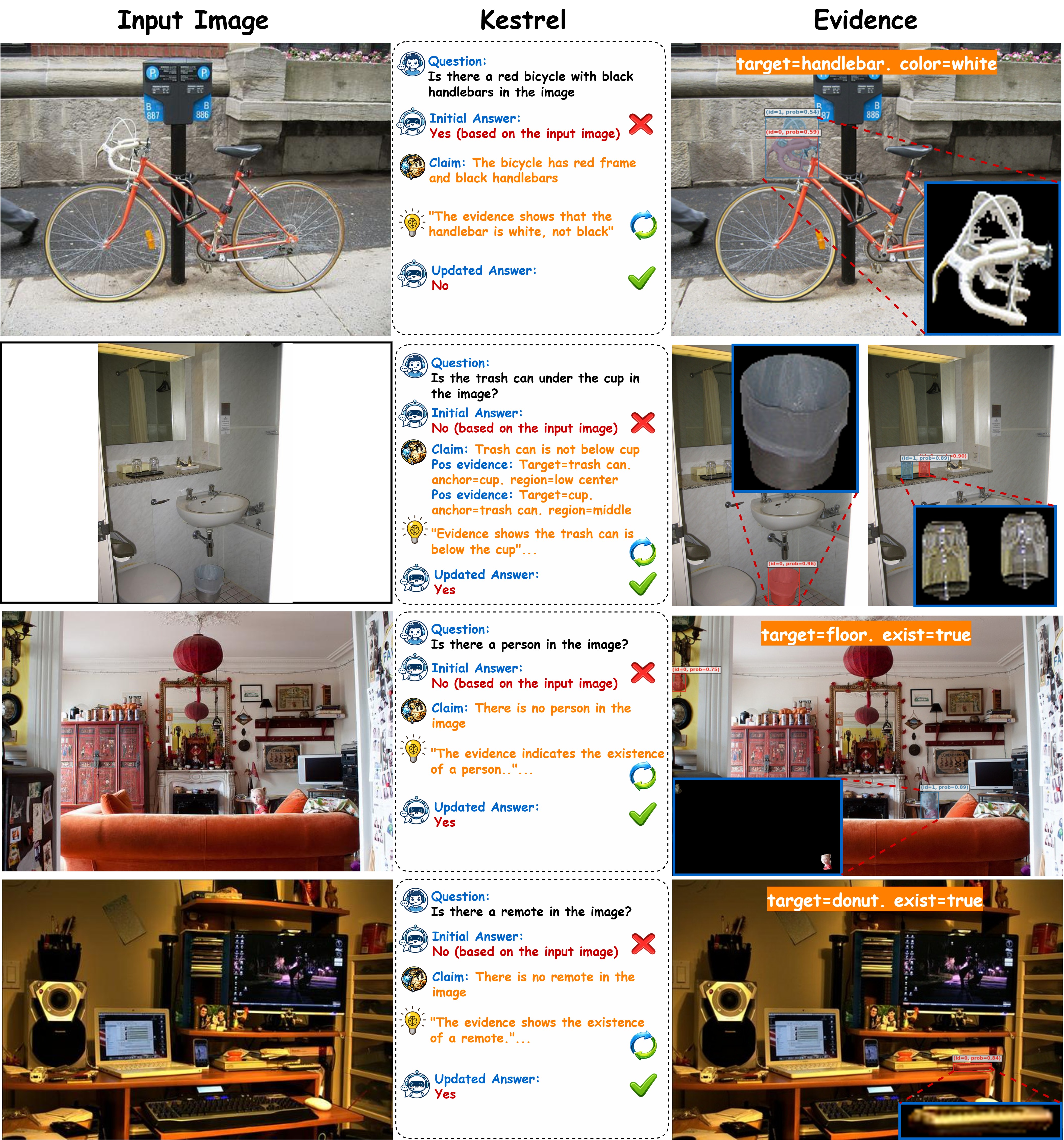}
    \vspace{-0.4cm}
     \caption{
     \textbf{Qualitative results of \name.} We compare the VQA responses from the regular baseline and our method based on InternVL-3.5. Zoom in for a better view.
     }
     \label{fig_app_q1}
   \end{figure*}

Fig.~\ref{fig_app_q} and Fig.~\ref{fig_app_q1} present representative examples of \name’s evidence-grounded self-refinement process. In each case, Qwen3-VL~\cite{bai2025qwen3} and InternVL3.5~\cite{wang2025internvl3} initially produce an incorrect answer based on their first-pass visual impression, while \name revises the prediction after introducing an explicit claim and verifying it against grounded evidence. The examples cover several queries, showing that the proposed framework is effective across different claim types. 
These cases highlight two key properties of \name. First, the method can correct errors even when the initial response is confidently wrong, by converting free-form reasoning into claim-level verification. Second, the refinement is interpretable: each answer update is supported by explicit visual evidence, such as segmentation overlays, cropped regions, or structured text evidence describing existence or position. Overall, the qualitative results show that \name improves faithfulness by grounding the refinement process in verifiable visual observations, rather than relying on unconstrained self-correction.

   \vspace{-0.4cm}

\name introduces a multi-stage, iterative inference pipeline that includes claim decomposition, external grounding, evidence construction, claim-level verification, and self-refinement. 
Although early stopping significantly reduces the number of refinement rounds in practice, the overall inference latency remains substantially higher than a single-pass LVLM inference due to the additional tool calls and verification steps.
This overhead may limit the applicability of the framework in latency-sensitive or large-scale deployment scenarios.
In future work, we plan to explore more efficient strategies for evidence-driven verification, such as adaptive tool invocation based on uncertainty signals, prioritizing high-risk claims for grounding, and reusing intermediate evidence across iterations.
Such improvements could substantially reduce the computational overhead while preserving the benefits of explicit grounding and iterative verification.

\section{Limitations and Future Work}

\name introduces a multi-stage, iterative inference pipeline that includes claim decomposition, external grounding, evidence construction, claim-level verification, and self-refinement. 
Although early stopping significantly reduces the number of refinement rounds in practice, the overall inference latency remains substantially higher than a single-pass LVLM inference due to the additional tool calls and verification steps.
This overhead may limit the applicability of the framework in latency-sensitive or large-scale deployment scenarios.
In future work, we plan to explore more efficient strategies for evidence-driven verification, such as adaptive tool invocation based on uncertainty signals, prioritizing high-risk claims for grounding, and reusing intermediate evidence across iterations.
Such improvements could substantially reduce the computational overhead while preserving the benefits of explicit grounding and iterative verification.

\end{document}